\documentclass[a4paper, 12pt]{article}

\usepackage{amsmath, amsthm, amssymb}
\usepackage{setspace}
\usepackage{indentfirst}
\usepackage{vmargin}
\usepackage{pdflscape} 
\usepackage{longtable}
\usepackage{multirow}
\usepackage{multicol}
\usepackage{natbib}
\usepackage{tabularx}
\usepackage[format=plain,labelfont={bf},textfont=it]{caption}
\usepackage{booktabs}
\usepackage{url}
\usepackage{bm}
\usepackage[top=1.5in, bottom=1.5in, left=1.5in, right=1.5in]{geometry}
\usepackage{endnotes}
\usepackage{epsfig}
\usepackage{psfrag}
\usepackage{amsfonts}
\usepackage[T1]{fontenc}
\usepackage{color}
\usepackage{rotating}
\usepackage{longtable}
\usepackage{graphics}
\usepackage{morefloats}
\usepackage{mathrsfs}
\usepackage{subfig}
\usepackage{colortbl}
\usepackage{lscape}
\usepackage{pdflscape}
\usepackage{verbatim}
\usepackage{array}
\usepackage[dvipsnames]{xcolor}
\definecolor{LightCyan}{rgb}{0.88,1,1}

\newcommand{\plcat}[1]{\textsf{#1}}
\newcommand{\plcon}[1]{\textbf{#1}}
\newcommand{\plmod}[1]{\texttt{#1}}
\newcommand{\plact}[1]{\textsc{#1}}
\newcommand{\pl}{\textsc{Plover}}
\newcommand{\pc}{\textsc{Polecat}}
\newcommand{\cm}{\textsc{Cameo}}
\newcommand{\cg}{\texttt{Chat\textsc{gpt}}} 
\newcommand{\pr}{\textrm{Prodigy}}
\newcommand{\soft}[1]{\texttt{#1}}  

\def\p3s{\phantom{xxx}}
\setpapersize{USletter}
\setmarginsrb{1.0in}{1.0in}{1.0in}{0.40in}{0in}{0in}{0in}{0.5in}

\begin{document}
\date{\today }
\title{Creating Custom Event Data Without Dictionaries: A Bag-of-Tricks
\footnote{This research was sponsored by the Political Instability Task Force (PITF). The PITF is funded by the Central Intelligence Agency. The views expressed in this paper are the authors' alone and do not represent the views of the U.S.\ Government. Funding for \pl ~was initially provided in part by the U.S. National Science Foundation award SBE-1539302, ``RIDIR: Modernizing Political Event Data for Big Data Social Science Research''.   We both acknowledge and are deeply grateful for the work on the Leidos side of this development by John Berrie, Heather Carter, Kathleen Krupenvich, and Jevon Spivey. All errors and interpretations remain our own.}}

\date{\today}

\author{Andrew Halterman\footnote{Michigan State University, \texttt{ahalterman0@gmail.com}}, Philip A. Schrodt\footnote{Parus Analytics LLC, \texttt{schrodt735@gmail.com}}, Andreas Beger\footnote{Basil Analytics, \texttt{adbeger@gmail.com}}, \\  Benjamin E. Bagozzi\footnote{University of Delaware, \texttt{bagozzib@udel.edu}}, Grace I. Scarborough\footnote{Leidos, \texttt{grace.i.scarborough@leidos.com}}}

\maketitle

\begin{abstract}
\noindent

Event data, or structured records of ``who did what to whom'' that are automatically extracted from text, is an important source of data for scholars of international politics. The high cost of developing new event datasets, especially using automated systems that rely on hand-built dictionaries, means that most researchers draw on large, pre-existing datasets such as ICEWS rather than developing tailor-made event datasets optimized for their specific research question. This paper describes a ``bag of tricks'' for efficient, custom event data production, drawing on recent advances in natural language processing (NLP) that allow researchers to rapidly produce customized event datasets. The paper introduces techniques for training an event category classifier with active learning, identifying actors and the recipients of actions in text using large language models and standard machine learning classifiers and pretrained ``question-answering'' models from NLP, and resolving mentions of actors to their Wikipedia article to categorize them. We describe how these techniques produced the new \pc ~global event dataset that is intended to replace ICEWS, along with examples of how scholars can quickly produce smaller, custom event datasets. We publish example code and models to implement our new techniques.

\end{abstract}

{\normalsize \clearpage}

{\normalsize \pagebreak }

\clearpage \pagebreak \renewcommand{\thefigure}{\arabic{figure}} %
\setcounter{figure}{0} \renewcommand{\thepage}{\arabic{page}} %
\setcounter{page}{1} \pagestyle{plain} 

\doublespace

\section{Introduction}

Event data, or structured records of ``who did what to whom'' that are automatically extracted from text, are an important source of data for scholars of international politics. The high cost of developing new event data, especially using automated systems that rely on hand-built dictionaries, means that most scholars draw on large, pre-existing datasets such as ICEWS rather than developing tailor-made event datasets optimized for their specific research question. 

This paper describes a ``bag of tricks'' for efficient, custom event data production, drawing on both recent advances in natural language processing (NLP) that allow researchers to rapidly produce customized event datasets and current work under the sponsorship of the Political Instability Task Force (PITF), to create a system, NGEC, that implements the \pl~ event data ontology to create a new global near-real-time dataset, \pc, with global coverage for 2010 to the present.\footnote{\pl/\pc ~is described in detail in \citet{halterman_et_al2023plover}.} We introduce techniques for training an event category classifier with active learning; identifying actors and the recipients of actions in text using large language models, standard machine learning classifiers, and pretrained ``question-answering'' models from NLP; and resolving mentions of actors to their Wikipedia article in order to categorize them. We also conduct initial comparisons with the hand-coded BFRS dataset  of political violence in Pakistan from 1988 to November 2011 \citep{bueno2015measuring}, and conclude with various pragmatic operational issues involved in the development and deployment of the system. We release Python code to implement the attribute- and entity-based techniques we describe in the paper, provided one develops their own text corpus and annotations.\footnote{This code can be found at: \url{https://github.com/philip-schrodt/NGEC} and \url{https://github.com/ahalterman/NGEC}.}

\subsection{Overview of event extraction}

Producing an event record from text requires extracting information from a document, categorizing it, and combining it into a single event record. In our approach, event extraction involves the following steps:

\begin{enumerate}
    \item \textbf{Event classification}: Given a document, we want to identify the event types that are reported in the text. For example, we might identify that a document reports a \plcat{protest} and \plcat{mobilize} event.\footnote{Throughout the paper, we make reference to the \pl~ event ontology. Note, however, that each of the steps are applicable to alternative or custom event ontologies, as we show in the section on replicating the BFRS dataset.}
    \item \textbf{Sub-event (mode) classification}: Given a document and a set of detected events, we then want to identify sub-categories (``modes'') of the event. For example, for a document where our event classifier identifies a \plcat{protest} event, we then identify whether the protest event is a riot, demonstration, hunger strike, etc.
    \item \textbf{\plcon{Context} classification}: Researchers would often like to know the \plcon{context} in which an event is taking place. For example, the protest could be reported in a news article that discusses  political corruption, the environment, or a pro-democracy movement. Knowing the \plcon{context} of an event helps researchers understand its role or allows them to limit their analysis to the set of events occurring alongside a specific context.
    \item \textbf{Event attribute identification}: We conceptualize political events as having several \textit{attributes}, namely the event's
    \begin{itemize}
        \item \textbf{Actor}: Which person, organization, or entity is initiating or carrying out the event?
        \item \textbf{Recipient}: Which entity is the event directed toward?
        \item \textbf{Location}: Where did the event take place?
        \item \textbf{Date}: When did the event take place?
    \end{itemize}
    We want to identify the phrases within the document that report this information for each event-mode combination.
    \item \textbf{Actor and location resolution}: Given the phrases extracted in the previous step, we would like to resolve different mentions of the same people, organizations, entities, and locations to a unique identifier in an external knowledge base. In practice, this involves identifying the Wikipedia article or entry in the Geonames gazetteer that correspond to each entity or location.
    \item \textbf{Entity categorization}: Finally, after resolving entities to their entry in Wikipedia, we would like to categorize entities based on their political role. For example, we might assign all military actors a \plcat{MIL} code to allow easier analysis of the data.
\end{enumerate}

The bulk of this paper describes new techniques for performing each of these tasks and compares the new techniques to previous approaches, which are generally dictionary-based. When used together in a complete pipeline, these steps will produce event data from news text. Most of the paper discusses the use of these steps to produce the new \pc ~dataset, but can be easily employed to produce custom data. Moreover, each step can be used separately to address specific research needs.\footnote{As this paper is references many existing datasets and previous coders, we include a glossary at the end of the paper.}

\subsection{Limitations of Dictionary-Based Coders}

Most existing, large-scale automated event datasets rely on keyword matching techniques to identify event types and assign labels to actors \citep{king2003automated, icews2015}. As a stylized example, a dictionary-based coder would compare a verb phrase such as ``threw weight behind'' against a dictionary that maps phrases to their corresponding event category (in the case of the \cm ~ontology used by ICEWS, Phoenix, and others, to an instance of \plcat{diplomatic cooperation}). Actors are resolved in a similar way, mapping generic terms (``soldiers'') or specific names (``Stanley McChrystal'') to a \plact{mil}[itary] role. Thus, most event coders have an event category dictionary that maps verb phrases to event types, and an actor dictionary that maps mentions of political actors to their country and category codes. Using dictionaries to identify events has several advantages: dictionaries are technically simple to implement, easy to interpret, and very fast to run over large corpora of text. Some event types and actors are easily identified using a small number of unique patterns, making dictionaries a useful tool for custom event datasets that are narrowly focused on a small number of actors and discrete events such as military offensives in Syria \citep{halterman2019linking}, rebel--government violence in Chechnya \citep{toft2015islamists}, or communal violence in India \citep{brathwaite2018measurement}.

Dictionary-based coders have a set of major limitations, however: They miss event descriptions and actors that are not present in the keyword list, they are expensive and labor intensive to develop, they require constant updates to reflect changing event types and political offices, and they struggle with interpreting words in context.

The ability to recognize all instances of an event or actor (i.e., recall) of dictionary methods is limited by their reliance on exact matching.\footnote{``Recall'' refers to the proportion of true events that a coder identifies. ``Precision'' is the proportion of coded events that are indeed true events.} If a verb phrase or actor is not included in the dictionaries, then it cannot be coded. \cite{althaus2018spatial} compare dictionary-based machine coded event to a gold standard, human-coded set of six event types related to protests and violence in Nigeria. The dictionary-based Phoenix event dataset has very poor recall compared to the gold standard, identifying between 0\% and 30\% of five event types when compared to human-coded data. This poor recall is often due to the lack of patterns in the event dictionary. For instance, the \cm ~open source verb dictionaries, the basis for the Phoenix datasets, contains no verb patterns at all for consequential event types such as \plcat{demand policy change}, \plcat{sexually assault}, \plcat{carry out roadside bombing}, and only a small number of patterns for a much larger set of event types.\footnote{For an analysis of the CAMEO verb dictionary's gaps see: \url{https://andrewhalterman.com/post/cameo-dictionary-coverage/}.} 

Because high-recall event coding requires very detailed dictionaries, developing dictionaries is costly and labor intensive. The difficulty grows for datasets that cover multiple event types across many contexts. The \cm ~verb dictionaries were under development from 1993 through 2016 and still have relatively low recall for identifying events \citep{althaus2018spatial}. \cite{halterman2018creating} report that translating 9,000 patterns from the English-language \cm ~verb dictionary into Arabic took a team of 15 Arabic speaking students a total of 750 hours of work. Manually adding other patterns and creating an actor dictionary took hundreds of hours more, even with advances in partially automating the task \citep{liang2018new}. The scale of effort involved in creating event data dictionaries makes any new project to create dictionaries from scratch a daunting one.

Finally, keyword-matching techniques struggle to interpret words in context. For instance, ``rolled into town'' has a very different meaning depending on whether it is preceded by ``an aid convoy'' or ``an armored column''. Simple text matching techniques are unable to incorporate the context of the sentence to correctly code ambiguous phrases. To avoid the costs associated with dictionary-based coders, we introduce a new, general, end-to-end event coder based on statistical classifiers trained on a new set of labeled data.

\section{Creating Event Data Without Dictionaries}

\begin{figure}
    \centering
    \includegraphics[width=\textwidth]{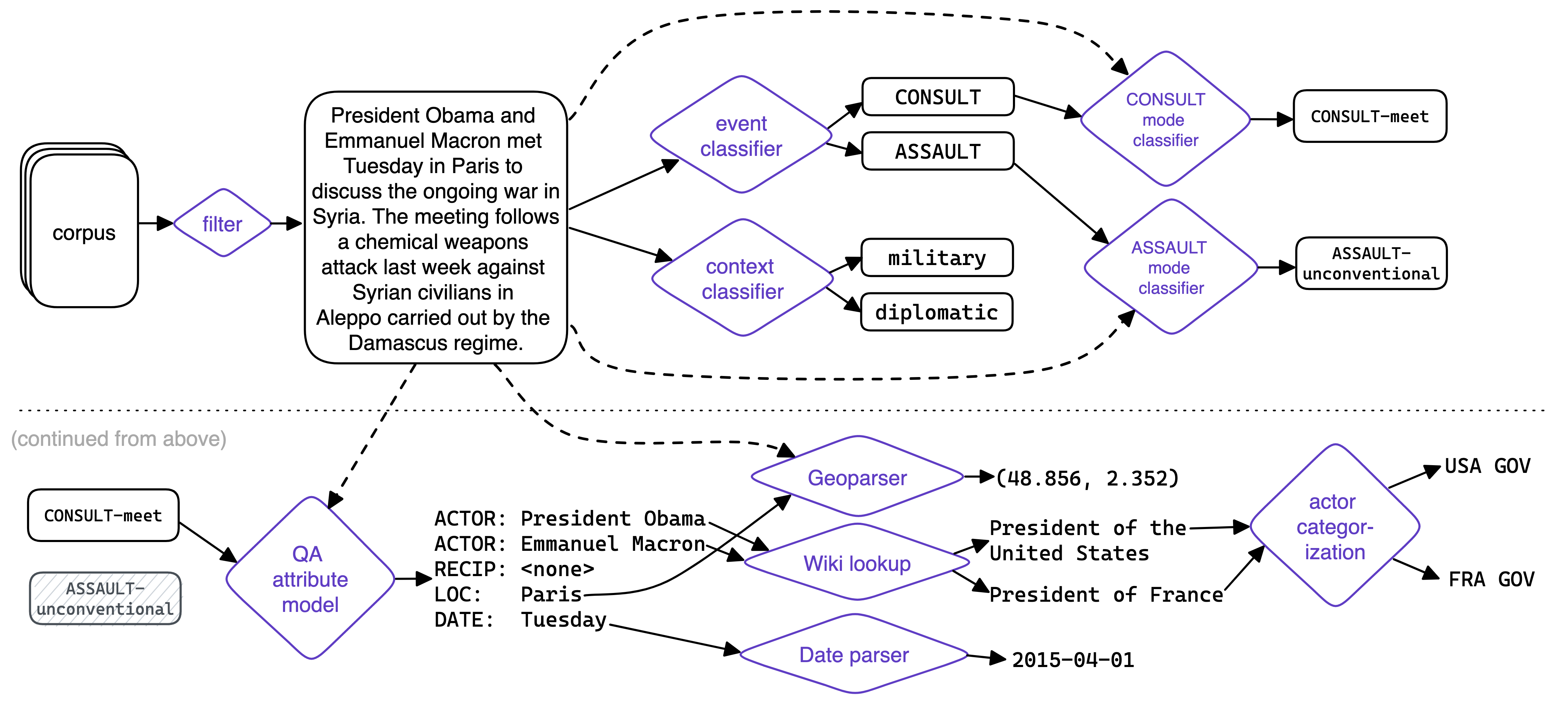}
    \caption{An overview of the entire event coding pipeline. Model steps are shown in blue and example outputs are shown in \texttt{fixed width} font. For space reasons, the pipeline is broken into two lines and the full coding of the \plcat{ASSAULT-unconventional} event is omitted. Dashed lines from the document indicate processing steps that use the entire document.}
    \label{fig:pipeline}
\end{figure}

The following sections describe our new techniques for performing each of the steps in coding event data from text: identifying events and sub-event types, identifying the event's context, extracting the text that reports the entities involved in an event, resolving these entities to a unique identifier, and categorizing the entities in useful ways. 
Together, these steps comprise the ``New Generation Event Coder'' (NGEC). An overview of the entire pipeline is shown in Figure \ref{fig:pipeline}. We also provide research code to implement the attribute identification and actor resolution steps. Throughout these sections, we discuss the process of producing the \pc~ global event dataset in the \pl~ ontology, which is described more fully in \citet{halterman_et_al2023plover}. However, the steps in the coder can easily be adapted to other coding schemes.

\subsection{Event classification with large-language-model-based classifiers}

Our event data pipeline begins with identifying the events reported in a document. To do so, we employ a \textit{document-level} classifier, specifically a transformer model fine-tuned on a corpus of documents that we labeled with the event types reported in the text. As an example, consider the short document below:

\begin{quote}
    President Obama and Emmanuel Macron met Tuesday in Paris to discuss the ongoing war in Syria. The meeting follows a chemical weapons attack last week against Syrian civilians in Aleppo carried out by the Damascus regime.
\end{quote}

\noindent In the \pl ~event ontology, we would like to be able to determine from the first sentence a meeting (~\plcat{CONSULT}). From the second sentence, we would determine that an ``\plcat{ASSAULT}'' event occurred.

In contrast to previous coders, our pipeline operates at the document level, rather than  sentence-by-sentence. This has advantages in not returning multiple codings of the same event described across several sentences in the same news story and for identifying the event's attributes (described below).

To identify the event types and modes in a story, we use transformer-based large language models (LLMs). Large language models such as BERT \citep{devlin2019bert}, GPT \citep{radford2019language}, and their variants, have revolutionized NLP in recent years. They are trained on a large corpus of text with the objective of inferring a ``masked'' word. By doing so, they generate representations of documents that encode much of the semantic and syntactic information of the text in a way that is useful for downstream tasks. These generic models can then be fine-tuned on a much smaller set of training cases specific to the problem domain, such as document classification, named entity recognition, document similarity, and question-answering models (all of which we employ in our pipeline). LLM's efficient transformer architecture \citep{vaswani2017attention} allows them to be trained on much larger corpora than previous models, such as LSTMs, and make them more efficient in practical applications.

BERT \citep{devlin2019bert} is a very widely used transformer model first published in 2019. The model was trained on a corpus of 3.4 billion words taken from English language Wikipedia and a collection of free novels by unpublished authors.\footnote{Training BERT required about 4 days of computation on a dedicated system with 64 GPUs, at an estimated cost of about \$7000, which underscores the usefulness of re-using existing models.} BERT is relatively slow to train and run, so we opt for distilBERT \citep{sanh2019distilbert}, a smaller version of BERT that runs much faster with only a modest reduction in accuracy.

The fact that transformers are pretrained on large amounts of text before fine tuning provides two obvious advantages over dictionary-based systems, and one unexpected advantage. First, by virtue of having seen distinct words in very similar contexts, they can easily infer synonyms and alternative phrasings of the same concept, whereas in dictionary-based approaches these have to be explicitly specified. Consequently, many dictionary patterns are very brittle, where changes in phrasing that would not even be noticed by a human reader cause a pattern to fail to match. This is particularly an issue for texts that are machine-translated or written by non-native speakers, both of which will tend to use words that are literally correct but would not be the obvious choice of a native speaker.

Second, these models look at words in context and thus can disambiguate multiple meanings of a word. For example, the phrase ``rolled into” has different meanings if the context is ``tanks'' (=take territory) or ``an aid convoy'' (=provide aid).
In dictionary-based systems, competing meanings must be disambiguated using a small number of usually proximate words, and are often brittle to small chances in phrasing. In contrast, the errors made by transformer models are often more similar to those made my humans. For instance, the event type \plcat{MOBILIZE} is found only rarely in the ICEWS data, but shows up fairly frequently in our \pc ~dataset, typically in conjunction with a use of force such as \plcat{PROTEST, COERCE}, or \plcat{ASSAULT}. Taking the \plcat{PROTEST} example, a text describing police dispersing protesters implicitly assumes that if the police were there, they must have first been mobilized. A dictionary-based system, in contrast, would require an explicit mention of the mobilization, and most articles would not do this, since the reader is assumed to know that for police to be at a protest, they first need to be mobilized.

Machine learning classification of events also provides another advantage over dictionary-based methods. Dictionary-based systems do a binary classification: a pattern fits the text or it doesn't. Machine-learning systems, in contrast, produce a classification score which is generally proportional to the likelihood that the classification is correct. This means the classifications can be calibrated for different levels of false positives and false negatives. We have been experimenting extensively with these as our current systems tend to over-classify, and so while we store all cases with a positive score, we only report those with scores above the 90-percentile threshold.

\subsection{The \plcat{event} category classifier}

The \plcat{event} category classifier is at the heart of the system. The \plcat{event} categories that are assigned to a text determine subsequent \plmod{mode} and entity processing, as well as whether the story is considered to contain events at all. Because of the importance of getting these initial steps correct, we use a  transformer-based model, which we found more accurate but more computationally expensive than other classifiers we considered.

\subsubsection{Discussion of 512-token limit in NGEC processing}\label{sec:512}

While our coder operates at the document rather than sentence level, a technical limitation in the classifier model we use limits us in some case to a shorter span of text than the full story. The BERT-family transformer models we use for the NGEC event coding have a 512 ``token'' limit on the amount of text they can work with. The WordPiece tokenizer \citep{devlin2019bert} that most transformer models use splits text into words, numbers, punctuation symbols, and, crucially, ``subword pieces'' to handle rare words and words not encountered in training. For example, the phrase ``the cities of Kyiv and Ulaanbaatar.'' would generate 11 tokens: [``the'',  ``cities'',  ``of'' ``kyiv'', ``and'',  ``ul'', ``\#\#aan:'', ``\#\#ba'', ``\#\#ata'', ``\#\#r'', ``.''], where the ``\#\#'' symbol indicates a split within a word.  Consequently, depending on the amount of punctuation and the number of rare words or proper names, 512 tokens will probably correspond to fewer than 450 words.
We have approximated the 512 word piece limit by using only the first 1024 characters in a text after first preprocessing to eliminate datelines and other systematically irrelevant material.\footnote{We are currently implementing this limit more precisely, both to get as much of the text as can be accommodated in 512 tokens, and also to get the exact coded text into the event output. In practice, the 512 word piece limit is not severely constraining. In a sample of 20,000 stories, 72\% of the articles were shorter than 512 tokens; in 50\% of the remaining cases additional characters roughly equal to 2/3rds to 100\% of the size (in characters) of the coded text is uncoded; the remainder of the cases have a very long tail.}

\subsubsection{Pre-processing filtering}\label{sec:preprocess}

While we originally expected that the transformer models would eliminate the need for pre-filtering of the text content, this proved not to be the cases, for at least two reasons. First, as we will be noting below, the transformers tend to over-generalize (a general issue with LLM models), and were particularly vulnerable given the relatively small, and not necessarily representative, number of training cases we had available. Second, the filters and pre-processing steps are relatively computationally inexpensive, and anything that can be eliminated at this stage does not incur the more costly steps of event and entity assignment.

While the source texts used in the project are relatively clean, the 20+ year backlog files involve tens of millions of texts from hundreds of sources accumulated under a wide variety of systems. As a consequence we have used various mundane filtering methods---largely string pattern matching and example-based SVMs---to remove datelines, editorial comments, and other extraneous information from the documents. We also detect and remove certain kinds of stories, including ``composite'' articles that report on multiple stories at once, financial stories, and stories that we detect as being primarily about crime. Further details on these steps are provided in the appendix. We also introduce a workaround for the problem of non-events, where an event is reported to have \textit{not} occurred. We use spaCy's part-of-speech tagger to identify instances of negated verbs and remove those sentences from the story. We hope to replace this system with a more sophisticated one that handles the negation more directly, returning the event with a ``negated'' tag or generating the appropriate event accounting for negation (e.g., ``will not sign the treaty'' $\rightarrow$ \plcat{REJECT}).

\subsubsection{Classification model estimation and selection}

Once the filtering is completed, the model estimation and classification is quite straightforward. A formatting and preprocessing step puts the stories to be classified into the right format for training. We then fit a binary classifier for each event type using Huggingface's transfomers library. Specifically, we fine-tune the pretrained `distilbert-base-uncased' \citep{distilbert2019} model for 3 epochs on our labeled data with default hyperparameters.

Because of both randomized split-sample training---some train/test sets are more representative than others---and the stochastic nature of model fitting, we fit 8 models for each category and then select a single ``consensus model'' that comes closest to mirroring the assignment performance of the entire ensemble. That ``consensus'' model is then used in the actual classification.

All of the estimates use a ``balanced'' 1:1 ratio of positive:negative training cases, a decision that may or may not have important consequences since in no instances does a \plcat{category} actually occur with 50\% frequency.\footnote{More details on the annotation process are provided in section \ref{sec:annotation}.}

\subsection{Mode and Context classification}

After identifying the events in an article, we  apply two additional classification steps to identify the \plmod{mode}  provided the assigned event \plcat{category} has modes, and \plcon{contexts} if applicable, which provides the ``why'' of the event. The \pl ~event ontology conceptualizes \plmod{modes} as the ``how'' for each event. For instance, after identifying an \plcat{ACCUSE} event, we then want to know the form the accusation took (allegation, disapproval, or investigation). To do so, we apply a mode-specific binary classifier for each \plmod{mode} that falls within an event that we identified in the story in the previous step. Similarly, binary classifiers for all of the \plcon{contexts} are applied to all texts that generated at least one event; \plcon{contexts} are independent of the event \plcat{category}.\footnote{The nature of our annotated data necessitates binary classifiers, rather than multi-class classifiers. See section \ref{sec:annotation} for details.} The development of these classifiers generally followed the same pattern as that of the event classifier, except that we found SVMs, which run in far less time than transformers and do not require a GPU, were sufficient and consequently the implementation and coding could be done much more quickly.

As with the event categories, we estimated a set of models---in the most recent iteration, 16, which takes about half an hour for \plmod{modes} and proportionately less for contexts, and selected the model with the best F1 score less than 1.0.\footnote{This last decision was made over concerns of over-fitting in an excessively easy random train/test split: this may or may not be a valid concern and could be revisited in the future} With very few exceptions, the accuracy metrics are well above the 0.70 to 0.80 range that can be achieved by well-trained human coders working at a sustained rate, and the precision and recall metrics are usually reasonably well balanced. Tables \ref{tab:context_acc} and \ref{tab:mode_acc} show the performance of these classifiers, as well as the wide variation in the sizes of the training sets, which are quite large for common behaviors such as \plcon{economic}, \plcon{military}, and \plcon{diplomatic} or \plmod{COERCE-arrest}, \plmod{PROTEST-demo}, and \plmod{REQUEST-assist}, but very small for a number of other categories.\footnote{The small sizes in many of the ASSAULT \plmod{modes} are partly an artifact of the coding environment: this was done before we had the street-crime filter in place and the yield of valid cases was quite low, though some \plmod{modes} such as \plmod{ASSAULT-abduct}, and \plmod{ASSAULT-assassinate} are simply rare in the overall corpus.} Again, these are metrics from split-sample testing, not against a large independent validation set, per Section \ref{sec:bench}, but nothing in these tables suggests the methodology is wildly inadequate.

\subsubsection{Post-processing events, modes, and contexts}

Finally, during the initial deployment of our coder, we noticed a high number of false positive codings for modes and contexts. While we hope to eliminate these with better training data and models, for now, we eliminate some mode and context labels using a secondary keyword filter. This is discussed further in Section \ref{sec:calib}.


\subsection{Identifying Event Attributes with Question-Answering Models}

After identifying  an event and its mode, we identify the actors involved in the event. The \pl ~ontology defines the following ``attributes'' for an event:

\begin{itemize}
    \item \textsf{ACTOR}: The entity carrying out the event.
    \item \textsf{RECIPIENT}: The entity that is the recipient, target, victim, or beneficiary of the event, depending on the event type.
    \item \textsf{LOCATION}: Where the event is reported to occur
    \item \textsf{DATE}: When the event is reported to have occurred.
\end{itemize}

\begin{figure}
    \centering
    \includegraphics[width=0.6\textwidth]{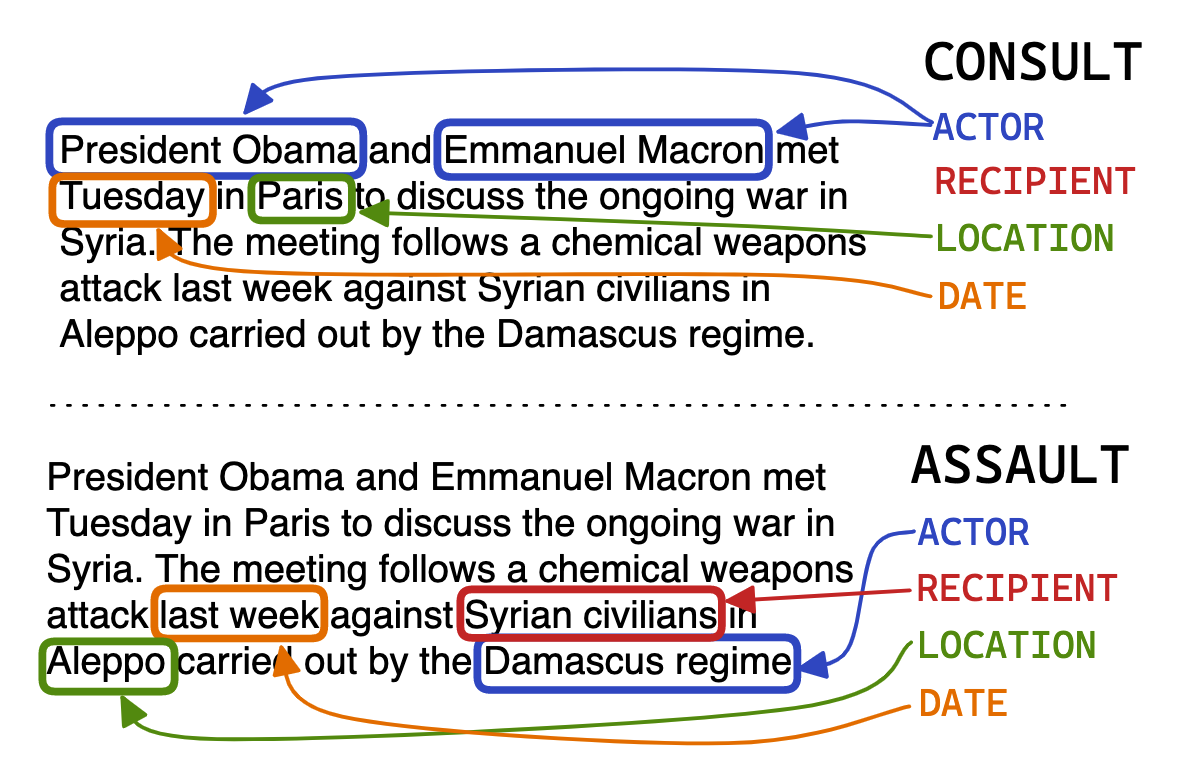}
    \caption{Example attribute identification. Note that the document contains two separate events (\plcat{CONSULT} and \plcat{ASSAULT}), each with its own attributes. The \plcat{CONSULT} event has two actors and no recipient, which is allowed for certain events by the \pl ~ontology.}
    \label{fig:attributes}
\end{figure}

Specifically, we want to extract the phrases of text that report the actor, recipient, location, and date of an event. Figure \ref{fig:attributes} shows an example of our desired output. Note that one document can report multiple events, and we extract the text that corresponds to each attribute for each event.

To identify each attribute's span of text, we employ an extractive question-answering (QA) framework. Extractive question-answering models in NLP take in (1) a passage of text and (2) a question written in natural language and return the span of text from the passage that best answers the question. We write a set of questions tailored to each event-\plmod{mode} combination to ask about its actor, recipient, location, and date. Table \ref{tab:event_mode_qs} shows example questions for two \plmod{modes} of \plcat{PROTEST}. For some event-modes, we write multiple questions to better elicit the correct span. We also write questions that use answers from previous questions, which greatly increases the coding time but also noticeably improves the accuracy,  especially for identifying recipients.

 Note that up to this point in the pipeline, our event and \plmod{mode} classifiers have identified the existence of an event--\plmod{mode} within the document, but not the precise passage of text that reports the event. One advantage of the QA approach to identifying attribute spans is that QA models can handle a common structure of news reporting, where an initial sentence describes an event in generic terms, and the sentences that follow provide more detail. The QA model generally prefers more specific mentions of entities, meaning that it can draw information from the second or third sentences of a story rather than returning generic actors from the first sentence. This is especially useful for geolocating events because more detailed locations are often not reported in the lede sentence. For example:

 \begin{quote}
      A bomb blast hit a religious school in northern Afghanistan on Wednesday, killing at least 10 students, a Taliban official said.
\\
The explosion went off at around the time of afternoon prayers at the Al Jihad Madrassa in Aybak, capital of Samangan province, a resident of the city who heard the explosion told The Associated Press.\footnote{Rahim Faiez, ``Taliban: 10 killed in bombing of Afghan religious school,'' \textit{AP}, 
November 30, 2022. https://apnews.com/article/afghanistan-bombings-taliban-3dbcb584545e8f8a2ac69185a4953a08}
 \end{quote}

Ideally, we would like to identify ``Aybak'' as the event location, rather than the more general ``northern Afghanistan'' in the first sentence.

\begin{table}[]
    \centering
    \renewcommand{\arraystretch}{1.2}
    \begin{tabular}{p{1.5in}p{1in}p{3.5in}}
    \toprule
    Event &  Attribute & Question \\
    \hline
PROTEST-demo & ACTOR	&  ``Who held a demonstration?'' \\
& ACTOR & ``Who held a demonstration against \{recip text\}?'' \\
& RECIP & ``Who was the target of the demonstration?'' \\
& RECIP & ``Who was \{actor text\}'s demonstration against?''  \\
& LOCATION & ``Where was the demonstration held?'' \\
PROTEST-riot & ACTOR & ``Who engaged in the riot?'' \\
 & ACTOR & ``Who rioted against \{recip text\}?'' \\
& RECIP & ``Who was the riot directed against?''	\\
& RECIP &``Who did \{actor text\} riot against?'' \\
& LOCATION & ``Where did the riot take place?'' \\
\bottomrule
    \end{tabular}
    \caption{Example questions event/\plmod{mode} combinations. Note that each event-\plmod{mode} can have multiple questions and questions can use answers from previous questions.}
    \label{tab:event_mode_qs}
\end{table}

We began with off-the-shelf transformer models fine-tuned on a question-answering dataset \citep{rajpurkar2018know}. However, when we evaluated the model's performance on a validation set with around 300 correct event attribute spans, the accuracy of the existing models was quite poor (F1 scores between 0.09 and 0.54, depending on the model and the attribute). We thus further fine-tuned a model on 4,823 newly collected annotations with our event-specific questions and answers. After further fine tuning on these new examples, we reach an F1 score of 0.86 for actors, 0.73 for recipients, and 0.91 for locations.\footnote{Our best performing model after fine-tuning is roberta-base-squad2, which was previously fine-tuned on SQuAD 2 and further fine-tuned on our new data.}

As an example of the entire process, consider the sentence ``Hindu nationalists rioted against Muslim shops in Delhi last week.'' We go through three rounds of questions to identify the event attributes:

\begin{enumerate}
\item a first pass with generic questions (``Who rioted against someone?'', ``Who was the target of the riot?'')
\item a second pass using detected actors from step 1: (``Who did \textit{Hindu nationalists} riot against?'')
\item a final pass using detected recipients from step 2: (``Who rioted against \textit{Muslim shops}?'')
\end{enumerate}

\begin{table}[]
    \centering
    \begin{tabular}{lll}
    \hline
    Span & Attribute & QA Model Score \\
   \hline
   \vspace{0.5em}
       A group of Hindu nationalists  & ACTOR & 0.433  \\
                        & ACTOR & 0.502 \\
                        & RECIP & 0.502 \\
                        \\
        Hindu nationalists & ACTOR & 0.452 \\
        \\
        Dehli & LOC & 0.939 \\
        \\
        Dehli last week & DATE & 0.751 \\
        \\
        Muslim shops & ACTOR & 0.131 \\
        & ACTOR & 0.131 \\
        & RECIP & 0.179 \\
        & RECIP & 0.755\\
        & RECIP & 0.131\\
        & RECIP & 0.103\\
        \hline
    \end{tabular}
    \caption{Example output from the QA attribute model. The QA scores are summed and spans are assigned to attributes using an optimal linear sum assignment method.}
    \label{tab:qa_output}
\end{table}

Because we ask multiple questions for each attribute and go through three rounds of questions, we have many possible span--attribute combinations, each with a confidence score returned by the QA model. We want to assign the best span to each attribute and do not want the same span to fill multiple attributes. We do so by summing the QA model scores for each span and using the Hungarian algorithm \citep{kuhn1955hungarian} to find the optimal span for each attribute.

\subsubsection{Broader literature on QA for event extraction}

Political scientists have long described event extraction as answering ``who did what to whom, when, and where?'' questions, so it's natural that question answering models would be useful for event extraction. Recent work in NLP has already applied QA models to event extraction. \citet{fitzgerald2018large} propose doing semantic role labeling, a similar task to event extraction, using QA models. More recently,  \citet{du2020event} conceptualize event extraction in a question-answering framework. Future work could draw more heavily on other contextual information present in an article. Unfamiliar political actors are generally introduced with a short description of their political role. Due to the difficulty in identifying these spans, the current coder does not attempt to identify actors' appositives or post-modifiers, but ongoing work on NLP could be a fruitful starting point \citep{kang2019pomo}.


\subsection{Resolving Entities to Wikipedia and Geonames with Neural Similarity}

After the previous steps, we now have events that have been identified from the text, along with the phrases that correspond to the actor and recipient for the action. Returning to our running example, we want to resolve the extracted entity mentions (``President Obama'', ``Damascus regime'') to a unique identifier, so  different mentions of the same person (e.g. ``President Obama'', ``Barry Obama'') are resolved to a single, canonical name (``Barack Obama''). This is usually referred to in the NLP literature as entity resolution. Performing this step correctly can already greatly reduce the number of unique actor strings in our data and makes it much easier to study individual people or organizations. We use Wikipedia as our external knowledge base and source of canonical entity names. For proper names, we use a process described below. Generic terms, such as ``Syrian civilians'' are not linked to Wikipedia and the process for handling these entities is described in section \ref{sec:entity-categorization}.

\subsubsection{Previous Work on Entity Resolution}

Previous event coders, such as the ICEWS coder or PETRARCH2, did not return a single canonical name for each detected entity. Instead, they resolved actor mentions directly to a higher-level category such as ``government'' or ``business'' (described below). This makes it impossible to do more fine-grained analysis within a category, such as examining which actors within a country's military are interacting with each other, and consequently, this limits the usefulness of previous event datasets to study bureaucratic politics, deliberations within the legislature, interactions between different rebel groups in the same conflict, or comparative politics more broadly.

\subsubsection{Creating a Custom Wikipedia Index}

We opt to create a local, offline version of Wikipedia rather than querying wikipedia.org directly.\footnote{Code to setup the offline Wikipedia index is located here: \url{https://github.com/ahalterman/NGEC/tree/main/setup/wiki}} Setting up our own index avoids the rate limitations that Wikipedia imposes, while allowing us to write custom search queries rather than relying on the built-in Wikipedia search bar functionality. This also allows our pipeline to be run without any external network connection, which is necessary in some situations.  

We download a dump of all English-language Wikipedia articles, and begin by extracting all redirects from within the Wikipedia dump. Redirects are hardcoded links from an alternative version of an entity name to its canonical Wikipedia article. For example, the Wikipedia page for the Islamic State has over 100 redirects, including many of the terms used in news stories, such as ``ISIL'', ``The Islamic State of Iraq and Syria'',  and ``Da‘ish''. After obtaining all the redirects, we parse each article into a simplified format containing the title of the page, any redirects to the page, alternative names automatically extracted from the first paragraph, the page categories, and Wikipedia's one-sentence summary of the article. We also extract and parse the infobox, if present, which includes information such as the office names and dates held for people holding political office. For space reasons, we retrain the introductory paragraph of the article but discard the rest of the article text. We skip category and disambiguation pages. 

We load the formatted articles into an Elasticsearch index that allows us to efficiently query the articles by any field. In total, after eliminating redirect, disambiguation, and category pages, we have 7,093,722 articles stored in our offline version of Wikipedia, out of 22,694,085 total pages.

\subsubsection{Neural similarity and Wikipedia-based actor coding}

To query the Wikipedia index with the actor text we would like to resolve, we begin with a fuzzy search over the Wikipedia index, searching within the title, alternative names, and redirects. We use a fuzzy search to better handle alternative transliterations of names, nicknames, or partial names.\footnote{Code for this process is available at \url{https://github.com/ahalterman/NGEC/blob/main/NGEC/actor_resolution.py}} As a result, we usually obtain many possible matches from Wikipedia. To select the best result, we use a combination of heuristic rules and a neural similarity approach. First, we check to see if the actor text has a single exact match for any of the article titles or a redirect and return it if so. If multiple Wikipedia articles have a match, we then employ a neural similarity approach, where we embed both the query term and information from the Wikipedia article into a fixed dimension encoder and compute the similarity between them. To train our model, we use the Wikipedia redirect data we collected earlier. From the redirect dictionary, we can construct pairs of examples, where either the pair is a true redirect (``Joseph Biden'' $\rightarrow$ ``Joe Biden'') or a false redirect (``Joseph Biden'' $\rightarrow$ ``Jill Biden''). When constructing the negative pairs, we sample page names that have high similarity but are incorrect to increase the difficulty of the task and ensure a better model. We fit a neural model with a constrastive learning objective \citep{chopra2005learning} to create embeddings for each page name that are similar to true redirects and distant from false redirects. Using a neural similarity approach allows us to identify different spellings of names, as well as handling query names that include components that may not be in the Wikipedia title, including titles, patronymics, middle names, or initials. 

At the end of this process, we ideally have a canonical Wikipedia identifier for each person or organization reported in the event. We evaluate our method on a dataset of named entities and their Wikipedia pages provided by \citet{hoffart2011robust}. When the true Wikipedia article is present in our results, we identify the correct article 84.0\% of the time. \citet{hoffart2011robust} do not report raw accuracy numbers, but report a precision-recall curve with an approximate value of 0.85 precision and 0.70 recall for their best method.

\subsubsection{Geolocation and Date Resolution}

We employ a different process for resolving place name mentions to a canonical entry in an external knowledge base. We use Geonames \citep{wick2011geonames} as this external knowledge base. The geolocation process is more fully described in \citet{halterman2023mordecai}, but we provide a brief summary here. We use spaCy's named entity recognition system \citep{honnibal2017spacy} to identify all locations in the news story.\footnote{Specifically, we use spaCy 3.4 with the transformer-based \texttt{en\_core\_web\_trf} model and identify all entities with a ``geopolitical entity' or ``location'' tag.} For each place name, we query Geonames, which is stored in an offline Elasticsearch index. When multiple matches occur, we select the best one using a model that takes into account the embedding of the placename, the other place names in the document, the content of the document itself, the similarity between the original place name and the alternative names reported in Geonames, and the predicted feature code and country of the extracted place name. The neural model is trained on several thousand annotated examples to rank the results and return the top match and its confidence. To determine which of the locations in the story is the event location for each event \citep{halterman2018linking}, we identify the geolocated place name that overlaps the location that the question-answering model identifies as the event's location. This results in a conservative event geocoding, because we only identify a geolocated entity as the event location if the entity overlaps the location answer from the previous QA step.

To identify the date on which an event occurred, we use the span identified by the QA model and use a rule-based system and the article's publication date to determine the event's date.\footnote{Specifically, we use Python's \texttt{dateparser} library.} By identifying that an event occurred on a previous date (``Tuesday'', ``last week'') and working backward from the story's publication date, we can get a more accurate sense of when events occurred. Previous event data systems generally used the story's date of publication as the event's date. However, many events in a story occur the day before publication or earlier. The problem is compounded by coding events from the full story, which often reports historical events for context at the end of the article \citep{althaus2022total}. By resolving phrases such as ``last November'' or ``in 2018'' to a date in the past, we help reduce historical events being assigned to the present.

\subsubsection{Limitations and Future Work}

Resolving entity mentions to their Wikipedia entries is a challenging and longstanding task in NLP \citep{hachey2013evaluating} and future work could improve the accuracy of our approach. One major limitation of our current approach is that we do not use the broader  story in resolving entities. For example, a mention of the ``Chamber of Deputies'' might not be resolvable to the correct country's legislature without including information from the broader story. Incorporating this contextual information would require changes to the model, along with additional training data.


\subsection{Entity Categorization}
\label{sec:entity-categorization}

As political scientists, we generally want to aggregate actors into broader categories of actors, rather than working with the (resolved) names of actors directly. For instance, we may want to group all of the legislators in a country, along with mentions of the legislature itself and its committees, into a single ``legislative'' actor for analysis. This second step is less common in the NLP literature and we refer to it as entity categorization. Table \ref{tab:actor_res_example} shows an example.

\begin{table}
\begin{center}
\begin{tabular}{lll}
    \textbf{Raw text} & \textbf{Resolved actor} & \textbf{Coded actor} \\
     President Obama & \texttt{Barack\_Obama} & USA GOV (in 2009-2017)\\
      Pentagon officials & \texttt{the\_Pentagon} & USA MIL\\
      Syrian civilians & \texttt{--} & SYR CVL\\ 
     Damascus regime & \texttt{--} & SYR GOV\\ 
\end{tabular}
\end{center}
\caption{Example of entity resolution and actor coding. Proper names of people and organizations are resolved to a canonical form (e.g., their Wikipedia article title), while generic terms do not have a corresponding Wikipedia article. The actor mentions are then further aggregated into a high-level code such as GOV, MIL, or CVL.}
\label{tab:actor_res_example}
\end{table}

\subsubsection{Previous Work on Categorizing Actors}

Most previous automated event data systems have used dictionaries to code actors, similarly to how they used them to code event types. A dictionary-based actor coding system compares an extracted actor to a large list of pre-specified actor patterns and returns the code for the matching pattern. The drawbacks of using a dictionary-based system were described above, but they are particularly acute for actor coding given the large number of possible patterns. For example, a simple string matching approach would fail to code ``demonstrators'' if it was not present in the dictionary. The problem becomes even more acute for proper names--``President Obama'' would not be coded if the only pattern in the actor dictionary was ``Barack Obama''. Furthermore, proper name dictionaries also require a date-based mechanism for returning the appropriate role as someone changes offices. Some techniques exist for automatically updating dictionaries (e.g., \citet{solaimani2017repair, radford2019automated}) but this process still involves significant ongoing manual labor to maintain a comprehensive set of actors and their associated codes.

\subsubsection{Neural similarity actor coding}

We employ a neural similarity model to assign codes to extracted actors. After the steps above, we have two types of actor mentions: generic actors such as ``Syrian villagers'' or ``government officials'', and proper names, such as ``Joe Biden'' that are now linked to their Wikipedia pages. In both cases, we would like to assign a broader actor category code to each.

We begin with a file of generic descriptions, offices, and titles, each with their associated PLOVER actor code. For example:

\begin{verbatim}
VILLAGE [~CVL]
DEPARTMENT_OF_AGRICULTURE [~GOVAGR]
DEFENSE_MINISTER [~GOVMIL]
REBEL_LEADER [~REB]
INTELLIGENCE_SERVICE [~SPY]
\end{verbatim}

We drew this list from earlier work \citep{schrodt2009tabari, norris2017petrarch2}, making some additions and changing the codes to reflect the new \pl ~actor ontology. Previous dictionary-based event coders required an exact match between an entry and a span of text to generate a coding. Thus, the existing file contains around 2,000 separate entries, including alternative spellings and slight variations (e.g. ``Defen\textit{s}e'' vs. ``Defen\textit{c}e'', ``Minister \textit{for}'' vs. ``Minister \textit{of}''). Despite the size of this file, dictionary-based systems still had problems with recall (false negatives). Instead of exact matching, we implement a neural similarity model to handle alternative terms and spellings. We generate an embedding for each entry in the file using a sentence tranformers model \citep{reimers-2019-sentence-bert} that produces a fixed-length embedding that performs well on neural similarity tasks.

For generic actors such as ``Syrian civilians'', we first detect any mention of a country or demonym to determine the country and remove it from the span. Then, we embed the remaining span using the same neural similarity model and compute the cosine similarity between the actor embedding and each of the embeddings in the file to identify the closest match. If the similarity is above a set threshold, we return the entry's code as the actor code for the span.

The process is similar for named entities that we link to Wikipedia. Once we have obtained a named actor's Wikipedia page, we extract components of the page that describe who the actor is. For officeholders who have an information box on the side of their Wikipedia page, we extract the job titles that correspond with the date of the news story.\footnote{Handling actors' changing roles over time is a critical component of an event coder intended for use on historical text.} For other types of actors, such as organizations, non-office holding people, or settlements, we obtain information from the page's single sentence summary, the information box type, or, as a last resort, from the introductory paragraph. Once we have an office title or short description for the actor, we then conduct the same neural similarity comparison described above, and return the code that best matches the person or organization's description on Wikipedia.


\section{Comparing to the BFRS Political Violence in Pakistan Dataset}

We now turn to a discussion of how our pipeline can produce custom event data using a different event ontology from the \pl ~ontology we use to code the \pc ~dataset.
While many researchers, especially those conducting cross-national or international relations research, want an off-the-shelf global event dataset of the kind we introduce in \citet{halterman_et_al2023plover}, other researchers have specific needs that are best addressed using custom event datasets. One of the major benefits of our dictionary-free pipeline is that it allows researchers to easily modify the pipeline to produce custom event dataset, including datasets that have different coding ontologies from \pl. This kind of modification is almost insurmountably difficult with dictionary-based coders because the entire dictionary needs to be written from scratch.

The BFRS dataset is a hand-coded event dataset of political violence in Pakistan from 1988 to November 2011 \citep{bueno2015measuring}. The 29,970 events were hand-coded from \textit{The Dawn}, Pakistan's major English-language newspaper and include 12 event types (terrorism, riots, drone strikes, conventional attacks, guerilla attacks, etc.), along with information about the responsible party, the reported cause, whether police or military forces were involved, the number of people killed or injured, and the date and location of the event.

The BFRS dataset is a good case for studying the applicability of the tools we introduce here to datasets beyond the new global POLECAT data. The BFRS dataset uses a different conceptualization of events from the one that we use. Several of the events have either the actor or recipient built into the definition (e.g., guerilla attack on security forces, security force attack on non-state combatants, etc.), while also separately reporting a ``party responsible''. The ``party responsible'' information uses a different classification scheme than the \pl ~scheme. The dataset also goes beyond the \pl ~ontology to include other information about the event, such as the number of people killed or injured and whether the event was successfully carried out or intercepted. We show how our pipeline can be easily be modified to produce data in BFRS's format.

Rather than using the original news text from \textit{The Dawn}, which is difficult to obtain for the entire period of the BFRS dataset, we opt to use the short description of the event written by the annotators as our document. Doing so makes our task much easier, as the description includes the details of the event without extraneous information that may have been in the original article. This means that our evaluation numbers will overstate the performance of the models. Applying the models to the subset of the original text that is available is an important next step in validating our method.

\subsection{Event Classification}

To simulate a hand coding process, we randomly sample 1,000 labeled events from the BFRS dataset (around 3\% of the total). Our random sample includes relatively few examples of some rare event classes, such as threats of violence and police/military selective violence, but this matches the process of hand labeling documents, where sampling rare classes for annotation is a difficult problem \citep{miller2019active}. On these 1,000 sampled events, we fit an efficient classifier that fine tunes an existing transformer model. Specifically, we fit a SetFit model \citep{tunstall2022efficient} to fine-tune a sentence transformer model, \texttt{paraphrase-mpnet-base-v2} \citep{reimers-2019-sentence-bert}.\footnote{We use a similar sentence embedding model in the actor categorization step described above.} For each document in the training corpus, SetFit samples documents of the same and different classes and uses a contrastive learning objective to produce embeddings that are similar for sentences of the same class and dissimilar for other classes. SetFit's efficient learning allows it to reach nearly the same performance as models with billions of parameters that are trained on thousands of examples, using a model with hundreds of millions of parameters and only dozens of examples. Training the model took about 5 minutes on a GPU and outperforms a simple embed-regress model by around 3 percentage points. Figure \ref{fig:bfrs_event} shows our ability to identify the BFRS event types in the event descriptions in our BFRS validation set.

\begin{table}
    \renewcommand{\arraystretch}{1.2}

\begin{tabular}{p{3in}p{0.8in}p{0.8in}p{0.8in}p{0.8in}}
\toprule
                                                   Event Type  &    Precision  &  Recall  &  F1-score  & N \\
\hline
conventional attack on military\slash paramilitary\slash police\slash intelligence  &     0.46  &    0.40  &    0.43 &       77 \\
                                                       terrorism   &    0.77   &   0.76   &   0.77  &     406\\
                                 violent political demonstration   &    0.83   &   0.83   &   0.83  &     247\\
     military\slash paramilitary\slash police attack on non-state combatants   &    0.62   &   0.87   &   0.73  &      69\\
                                                   assassination   &    0.86   &   0.81   &   0.84  &     616\\
                                    assassination (drone attack)   &    1.00   &   0.86   &   0.92  &      14\\
                                                           other   &    0.21   &   0.30   &   0.25  &     129\\
                                                            riot   &    0.83   &   0.76   &   0.79  &     193\\
                                              threat of violence   &    0.18   &   0.25   &   0.21  &      12\\
                                                 attack on state   &    1.00   &   0.83   &   0.91  &      81\\
    guerilla attack on military\slash paramilitary\slash police\slash intelligence   &    0.48   &   0.57   &   0.52  &     107\\
                 military\slash paramilitary\slash police selective violence   &    0.62   &   0.29   &   0.39  &      45\\
\hline
                                                        accuracy   &           &          &   0.73  &    1996\\
                                                       macro avg   &    0.66   &   0.63   &   0.63  &    1996\\
                                                    weighted avg   &    0.75   &   0.73   &   0.73  &    1996\\
\bottomrule
\end{tabular}
\caption{Validation set accuracy (N=1,996) of our BFRS event classifier trained on 1,000 randomly sampled documents. The classifier is paraphrase-mpnet-base-v2 fine-tuned using SetFit.} 
\label{fig:bfrs_event}
\end{table}

For most events, the classification F1 score is acceptable (over 0.8). For others, including guerilla attacks on security forces, threats of violence, and the ``other'' category, further annotations would be required. Based on work developing the NGEC event coder, we would recommend an active learning approach to decide which documents to annotate, using the initial model to suggest examples where the existing model is least certain \citep{miller2019active}. 

\subsection{BFRS attribute identification}

While the BFRS event dataset does not record actors and recipients in the same way as the PLOVER ontology, it still reports the ``party responsible'' and the location of the event. We can apply our existing QA model with new questions that are tailored to the BFRS event definitions to obtain information on the location and party responsible. Writing new questions for the 12 event types is relatively easy for a researcher with some knowledge of subnational conflict and took less than an hour to do. Because our original QA-attribute model was trained on a diverse set of text and event types, it performs well even on the new BFRS event types. Table \ref{tab:bfrs_actor} shows the output of the QA-attribute model on four event descriptions taken from the BFRS dataset. Note that the model handles passive construction well, but sometimes misses some details of the actors, including an instance where it selects the generic ``gunmen'' instead of the more precise ``Baloch Liberation Army.''

\begin{table}
    \renewcommand{\arraystretch}{1.2}
\begin{center}
\begin{tabular}{p{1.2in}p{5in}}
\toprule
     Event Type & Event Description  \\
    \hline
    \vspace{-0.5em} \\
    
     conventional attack on security forces & three people, a \textbf{\color{BrickRed} security personnel} among them, were injured when \textbf{\color{Aquamarine} gunmen} riding a motorcycle opened fire on a security vehicle in \textbf{\color{OliveGreen} Khuzdar town}. Mirack Baloch, spokesman of the Baloch Liberation Army, claimed responsibility for the attack. \\
     drone attack & Seven \textbf{\color{BrickRed} suspected militants} were killed while another four were injured when a \textbf{\color{Aquamarine} US drone} fired missiles near \textbf{\color{OliveGreen} Angoor Adda in  South Waziristan}  \\
     violent political demonstration & \textbf{\color{Aquamarine} Students from local colleges} took out one protest demonstration while religious parties took out another demonstation. Both combined into one at \textbf{\color{OliveGreen} Jogi Chowk}. The protesters asked \textbf{\color{BrickRed} shopkeepers} to close their shops. One shopkeeper in Kariana Bazaar resisted and the protesters damaged his shop. They also set fire to tyres and took down film boards that were put up in the streets.  \\
     threat of violence & A letter was sent to \textbf{\color{BrickRed} shopkeepers} in \textbf{\color{OliveGreen} Ghafoor Market of Charsadda Bazaar} allegedly by the \textbf{\color{Aquamarine} local Taliban} warning them against dealing with women shoppers. \\
     guerilla attack on security forces & \textbf{\color{BrickRed} Three people} were killed while more than twenty others, including policemen, were injured when a \textbf{\color{Aquamarine} suicide bomber} exploded a pick-up truck near a police station in \textbf{\color{OliveGreen} Darra Pezu town in Lakki Marwat}. A police constable opened fire on the vehicle as it raced towards the police station, the bomber could not reach the walls of the station and blew up before it. Over thirty shops were razed to the ground and nearby houses were damaged. The walls and mosque of the police station were damaged as well. \\
    \bottomrule
\end{tabular}
\end{center}
\caption{Example descriptions of different event types from the BFRS dataset showing the automatically detected \textbf{\color{Aquamarine} ACTOR}, \textbf{\color{BrickRed} RECIPIENT}, and \textbf{\color{OliveGreen} LOCATION} attributes. Dates are rarely reported in the dataset's event descriptions field.} 
\label{tab:bfrs_actor}
\end{table}

\subsection{BFRS actor categorization}

Because the BFRS dataset categorizes actors in a different way than the PLOVER ontology, we cannot simply apply the existing file of generic roles and job titles. However, creating a new file that maps actor descriptions to the 15 actor categories is again straightforward. We take terms and examples from the BFRS codebook to create a new actor categorization file.


\subsection{BFRS geolocation}

Next, we evaluate our ability to recover the location of events reported in text. The BFRS dataset reports the location, tehsil (sub-district), district, and province for each event. Evaluating accuracy at the finest grained location is difficult because alternative spellings between the BFRS dataset and the Geonames gazetteer mean that we cannot directly compare the location names without hand validation. As an alternative, we assess province-level geocoding, which is shown in Table \ref{tab:bfrs_province_geo}. Because the event descriptions reported in the BFRS dataset do not always include the location information from the original article, many events cannot be geolocated. We exclude the 17,338 locations out of 29,969 total events that do not have detected locations in their description. As Table \ref{tab:bfrs_province_geo} shows, the geolocation process is accurate most of the time, but still makes errors at the provincial level. 

\begin{table}[]
    \centering
    \begin{tabular}{l|llllll}
    \hline
& \textit{Balochistan} & \textit{KPK} & \textit{Sindh} & \textit{Punjab} & \textit{AJK} & \textit{GB} \\
    \hline
Balochistan & \textbf{677} & 82 & 64 & 143 & 8 & 13 \\
KPK & 136 & \textbf{791} & 84 & 169 & 13 & 6 \\
Sindh & 542 & 481 & \textbf{3083} & 888 & 108 & 76 \\
Punjab & 189 & 550 & 325 & \textbf{1898} & 77 & 12 \\
AJK & 46 & 187 & 48 & 108 & \textbf{131} & 0 \\
GB & 3 & 3 & 3 & 7 & 0 & \textbf{29} \\
    \end{tabular}
    \caption{Province-level geolocation accuracy for the events in the BFRS dataset. True labels are shown in the row labels, predicted provinces are in italics as the column headers. Out of 29,969 events, 17,338 did not have a detected location in the description text and are excluded from this table.}
    \label{tab:bfrs_province_geo}
\end{table}

\section{Some operational considerations}



\subsection{Reduced labor and training}

Dedicated work on NGEC began in mid-August 2021 and all operational code had been delivered by mid-March 2022. In contrast, the four major dictionary-based coders which produced data used in academic publications—KEDS (ca. 1990-1994), VRA-Coder (ca. 1992-1996), TABARI (ca. 2000-2004) and ACCENT (ca. 2011-2015) each required about four years of development—generally first to get a working coding platform, then a long period of incrementally refining dictionaries to work with this—and except for KEDS, built on the customized dictionaries of earlier projects. In addition to these successful projects, probably twice as many projects, most recently a three-year, \$2-million NSF-sponsored effort, were never able to successfully deploy a new coding system.\footnote{There also have been successful projects that were, effectively, hybrids which extended existing systems: these would include Lockheed's JABARI, a Java-based extension of TABARI, the Open Event Data Alliance PETRARCH-1 and PETRARCH-2 coders—the latter a summer project by a single very bright undergraduate intern—which leveraged Stanford's CoreNLP parsers with TABARI dictionaries, and a variety of attempts, typically in pursuit of DARPA funding, by major software corporations such as IBM to modify large scale natural language processing systems developed for other purposes for event coding, but none of these had wide acceptance.}

The efficiency of NGEC is due primarily to the transformer revolution and the fact that most of the models we needed are available in well-documented open-source platforms such as HuggingFace (https://huggingface.co/).  The transformer models, in this and many other applications, have proven to be remarkably adaptable and robust so, for example, we can use variations on the same technology—and based on the same underlying pre-trained models—to deal with both events as a classification problem and entities as a QA problem. While less dramatic in terms of capabilities, the availability of open machine learning suites such as Python's \soft{sklearn} allowed us to rapidly experiment with, and then implement, the SVM classifiers.

We have also been able to leverage additional very large open source resources—notably the Wikipedia and Geonames databases—to solve the issue of keeping dictionaries complete and up-to-date that would otherwise require ongoing maintenance by trained staff. This is particularly relevant to actor and recipient information, which is constantly changing due to mortality, elections, internal changes in governments, and the emergence of new actors.

\subsection{Computational requirements}

The computational requirements are quite significant in this project. In particular, it was simply impractical to use large language models on the \plmod{modes} and \plcon{contexts} because of the number of required binary models,\footnote{For annotation efficiency, we only collected binary annotations on modes and contexts (e.g., does the story have a ``gender'' context?), not complete annotations for all modes and contexts. As a result, we cannot train a multi-class classifier without sophisticated techniques to handle missing labels.}  and even estimating the core models (which are based on multiple estimated models) for the event \plcat{categories} requires about 36 hours of run-time using a GPU on the Google Colaboratory system. Working without a GPU, which provides an increase of speed of 30x to 60x, is all but impossible, though it was possible to do the original proof-of-concept with a conventional personal computer. The attribute step is by far the most computationally costly for two reasons. First, we found the accuracy loss from using a smaller, more efficient model such as DistilBERT or TinyBERT, rather than the larger and slower RoBERTa model to be unacceptable. Second, we run the question-answer model three separate times, as discussed above, in order to use previous answers to improve our questions. We found this an important step in increasing the accuracy of recipient identification, but it increases our time by more than 3x.\footnote{Because we ask multiple questions in each round, we can end up in a combinatorial process of asking more questions each round, as we incorporate answers from the previous round into subsequent questions.}

These long processing times have limited the amount of experimentation that is possible, and whenever LLM-based components are added to improve performance, processing requirements increase significantly. Consequently where possible we've used conventional machine-learning classifiers such as SVMs, which can run our training corpora in tens of minutes rather than tens of hours. In contrast, processing time for dictionary-based systems such as \soft{TABARI, PETRARCH} and \soft{ACCENT}, while not trivial for large corpora, was never a significant constraint.

We see three potential mitigations to the slow processing time. The easiest approach is to use the increasing speed of GPUs and efficiencies in off-the-shelf models. The second approach would consist of applying better software engineering to our pipeline. Modern GPUs can efficiently batch several tasks together, but our batch size for the attribute model is limited by the number of questions we ask for each article. At the cost of much greater complexity, we could collect questions from multiple articles and batch them together, potentially lowering our computing time. Finally, access to more training data would allow us to use smaller, more efficient models. We benefit from the ability of larger language models to perform transfer learning to our domain. With more training data, however, we could more fully train a smaller language model to perform well on our task.

Our processing pipeline as a whole is easy to run in parallel, however. Each article's processing is independent of other articles', meaning that we can easily distribute the processing across multiple GPUs, or machines. 

\section{Future Work on Benchmarks and Evaluation Data}\label{sec:bench}

While obvious on reflection, it's worth noting that while dictionary-based models are largely opaque unless one has access to the dictionaries, which we do not have for event coding and many actors in \soft{ACCENT}, the models of any example-based system can be effectively reverse-engineered provided the basic features are known: one doesn't need the exact set of training cases, just a comparable set of inputs to match to a known set of outputs (in other words, unlike with a dictionary-based system, one doesn't need the \textit{exact} sentence describing the event, just a text that looks similar to an LLM: these comparable stories can be readily found just from the date and actors and located coded into the event record). These input-output pairs are available in the millions explicitly in the government iDATA which has the source texts, and can be easily inferred from the public data since with few exceptions it is easy to locate archived texts which correspond to a given event. While there are countless ways to implement neural networks, it is likely that these would converge to accuracy as good as, or better, than ours. 

Nonetheless, it would be useful to have a standard set of benchmark cases that could be used to evaluate systems. These provide metrics that academic computer scientists use to evaluate and, if successful, publish new enhancements. Computer science literature tends to be both timely and open access and is focused on beating the previous state-of-the-art on established benchmarks, so this essentially provides free and sometimes highly trained labor to explore possible modifications to the system. Since to date we've just used default values on the ``hyperparameters'' of the models---these are characteristics of the models which cannot be estimated from the data—there are presumably ways these can be improved. Those improvements are generally marginal—1s of percentages in accuracy—but are, nonetheless, improvements, and occasionally major breakthroughs will be found. The computer science community is also likely to experiment with new models and methodologies and could alert us to improvements.

Finally, there is interest in this topic within the computational linguistics community: in both 2020 and 2021 the Language Resources and Evaluation Conference had a well-attended (>100 participants) workshop on ``Automated Extraction of Socio-Political Events from News Sources''—which is to say event data—with quite a few participants and some modeling competitions. 

To fully exploit the current machine learning environment at all levels of the system, we ideally need a large and representative ``gold standard''\footnote{We'd be happy with a bronze standard\ldots} validation data set with complete coding (in contrast to the component-by-component selections in our various training sets) of a large number of cases out of a representative corpus.\footnote{This is also distinct from the [in]famous (but open source!) ``Lord of the Rings'' validation set used in the development of \soft{TABARI} and the \soft{PETRARCH} family of coders: these assured that various components of the parser-coder continued to work correctly, but did not systematically check the resulting data. BBN undoubtedly had something similar for \soft{ACCENT} but, like the \soft{ACCENT} verb patterns, this is not open.} If this were available, a long series of additional refinements and metrics would be open, notably:
\begin{itemize}
    \item experimentation on hyperparameters such as the positive/negative balance in training cases
    \item on-going assessment of dysfunctional cases in the training sets
    \item calibration of classification score levels
    \item estimation (and reduction) of the false negative level (note that this is not even known for ICEWS). See \citet{halterman2021corpus} for a discussion of the difficulties in estimating recall for event data.
\end{itemize}

More generally, if a reference set were available, we could also readily utilize a wide variety of open source tools readily accessible from the academic community and industry.

The issue, as always, is cost, as human-annotation is needed to at least review the cases. A project could, however, be distributed across a number of institutions (with, of course, near-real-time automated checks to ensure we are not seeing individual coder or institution effects, which in the absence of monitoring are almost inevitable and have, alas, only been discovered retrospectively in a number of widely-used human-coded data sets), possibly with a combination of projects using undergraduate and graduate coders, professional coders, and feedback from the user community. In an ideal world, no assessment of a coding, positive or negative, would ever go unrecorded, that is, when a user says ``This coding is the dumbest thing I've ever seen!'' that's critical and quite important data we would like to recover, though putting systems in place to do this is difficult.

As far as validation sets are concerned, the bigger the better, and the ideal answer to ``how much should this cost?'' is the dreaded ``How much do you have?'' But realistically, a combination of getting low-hanging fruit via the existing classifiers (or ensembles of standard classifiers, this output being generally less fatiguing to annotate as it is largely positive), a distributed effort across various research and NGO groups, and harvesting feedback from users, it is probably could be done with a relatively modest investment. The reference data would need to be free of intellectual property constraints---under US copyright law, it probably is already due to the ``fair use'' exceptions for research and the fact that the examples are an utterly minuscule proportion of the original corpus, but ``anonymizing'' transformations, as are currently common in public research data involving human subjects, or using the cases as input to a LLM-based system that transmogrifies into a synthetic cases whose effects on training are indistinguishable from the original (again, with a validation set, we can assess this.) That reference set would then effectively define \pl, and could be used by the conflict studies, computational linguistics, and machine learning communities for years to come: ``theories come and go, a good data set lasts forever.''

\section{Conclusion}

This work introduces a new end-to-end event coder, NGEC, that moves beyond the limitations of previous dictionary-based coders. It draws heavily on recent advances in NLP, including transformer models for classification, question-answering models for event attribute identification, and neural similarity techniques. We draw on existing resources, including Wikipedia and Geonames, to eliminate the need to maintain large sets of actor dictionaries. We have deployed the pipeline to produce \pc, a new global event dataset that is intended to replace the ICEWS dataset.

Each of the steps we describe here are modular and can be used on their own to address researchers' specific needs. For instance, our active learning annotation scheme and transformer-based classifiers are useful for researchers who want to perform document classification. Our geoparsing techniques are useful for researchers who need to extract and resolve place names from a document without performing full event extraction, or for researchers who have semi-structured data on locations and want to resolve them to coordinates. Our Wikipedia lookup process is broadly useful for researchers who have multiple mentions of political actors that they would like to resolve to a single canonical identity.

A main objective in developing the pipeline was to lower the cost for producing new, custom event datasets. Researchers are often interested in event types or actor categorizations that are not present in off-the-shelf datasets, but the costs involved in hand-coding event data or developing dictionary-based event coders are prohibitive for most projects. We demonstrate that our pipeline can easily be repurposed to code event data that uses event ontologies that differ from the \pl ~ontology that we employ. We hope that the techniques and code we implement here will dramatically lower the cost to creating new event data sets.

\clearpage
\newpage
\bibliographystyle{apsr}
\bibliography{bibliography}

\clearpage

\begin{appendix}

\section{Accuracy evaluation for event, mode, and context classifiers}
\renewcommand{\thefigure}{A\arabic{figure}}
\renewcommand{\thetable}{A\arabic{table}}
\setcounter{figure}{0}
\setcounter{table}{0}

\renewcommand*{\arraystretch}{1.4}

\begin{table}[]
\scriptsize
    \centering
    \begin{tabular}{lllll}
\hline
\plcat{AGREE} \\ 
Mk7  &  Train size: & 388  & Eval size: &  97 \\ 
Mk9 &  Train size: & 239  & Eval size: &  60 \\ 
   &  Acc &  Precision & Recall  &    F1 \\ 
Mk7 Mean & 0.7126  & 0.6408   & 0.7744   & 0.6950 \\ 
Mk7 StdDev & 0.0320  & 0.0574   & 0.0797   & 0.0121 \\ 
Mk9 Mean & 0.7167  & 0.7373   & 0.8816   & 0.7983 \\ 
Mk9 StdDev & 0.0408  & 0.0548   & 0.0721   & 0.0136 \\ 
\hline
\plcat{CONSULT} \\ 
Mk7 &  Train size: & 404  & Eval size: & 101 \\ 
Mk9 &  Train size: & 337  & Eval size: &  85 \\ 
   &  Acc &  Precision & Recall  &    F1 \\ 
Mk7 Mean & 0.8998  & 0.8920   & 0.9100   & 0.9004 \\ 
Mk7 StdDev & 0.0273  & 0.0435   & 0.0173   & 0.0253 \\ 
Mk9 Mean & 0.8603  & 0.8307   & 0.9884   & 0.9015 \\ 
Mk9 StdDev & 0.0592  & 0.0567   & 0.0129   & 0.0346 \\ 
\hline
\plcat{REJECT} \\ 
Mk7 Mean  &  Train size: & 373  & Eval size: &  94 \\ 
Mk9 &  Train size: & 247  & Eval size: &  62 \\ 
   &  Acc &  Precision & Recall  &    F1 \\ 
Mk7 Mean & 0.5425  & 0.5092   & 0.6477   & 0.5690 \\ 
Mk7 StdDev & 0.0213  & 0.0175   & 0.0612   & 0.0274 \\ 
Mk9 Mean & 0.4597  & 0.5245   & 0.5571   & 0.5305 \\ 
Mk9 StdDev & 0.0228  & 0.0244   & 0.1262   & 0.0604 \\ 
\hline
\plcat{THREATEN} \\ 
Mk7 & Train size: &  71  & Eval size: &  18 \\ 
Mk9 & Train size: &  40  & Eval size: &  11 \\ 
  &  Acc &  Precision & Recall  &    F1 \\ 
Mk7 Mean & 0.5417  & 0.5863   & 0.6000   & 0.5910\\ 
Mk7 StdDev & 0.0666  & 0.0551   & 0.0866   & 0.0625\\
Mk9 Mean & 0.4431  & 0.5456   & 0.7678   & 0.6345 \\ 
Mk9 StdDev & 0.0301  & 0.0192   & 0.0994   & 0.0361 \\ 
\hline
\plcat{ASSAULT}\\ 
Mk7 &  Train size: & 390  & Eval size: &  98 \\ 
Mk9 &  Train size: & 212  & Eval size: &  53 \\ 
   &  Acc &  Precision & Recall  &    F1 \\ 
Mk7 Mean & 0.7398  & 0.7081   & 0.7772   & 0.7383 \\ 
Mk7 StdDev & 0.0542  & 0.0677   & 0.0541   & 0.0447 \\ 
Mk9 Mean & 0.6250  & 0.6785   & 0.2881   & 0.3494 \\ 
Mk9 StdDev & 0.0476  & 0.1330   & 0.2227   & 0.2177 \\ 
\end{tabular}
    \caption{Comparative accuracy figures for the document-level event \plcat{category} classifier before (Mk7) and after (Mk9) filtering of irrelevant training texts and manual review of training cases. }
    \label{tab:category_acc}
\end{table}

\begin{table}[]
    \centering
    \begin{tabular}{lllllll}
\plcon{context}          & Train N & Test N & Accuracy &  Precision &  Recall &     F1   \\    
\hline
asylum           &   269   &    68  &  0.868  &   0.943  &  0.825  &  0.880  \\
corruption       &   301   &    76  &  0.802  &   0.765  &  0.929  &  0.839  \\
crime            &   258   &    65  &  0.831  &   0.774  &  0.857  &  0.814  \\
cyber            &   198   &    50  &  0.840  &   0.833  &  0.893  &  0.862  \\
diplomatic       &   190   &    48  &  0.729  &   0.833  &  0.758  &  0.794  \\
disasters        &   196   &    50  &  0.840  &   0.842  &  0.762  &  0.800  \\
economic         &   412   &   103  &  0.796  &   0.870  &  0.727  &  0.792  \\
election         &   306   &    77  &  0.727  &   0.745  &  0.796  &  0.769  \\
environment      &   294   &    74  &  0.865  &   0.897  &  0.854  &  0.875  \\
gender           &   204   &    52  &  0.769  &   0.926  &  0.714  &  0.807  \\
health           &    92   &    24  &  0.792  &   0.706  &  1.000  &  0.828  \\
inequality       &   301   &    76  &  0.790  &   0.784  &  0.784  &  0.784  \\
intelligence     &   233   &    59  &  0.780  &   0.788  &  0.813  &  0.800  \\
legal            &   297   &    75  &  0.733  &   0.725  &  0.763  &  0.744  \\
legislative      &   170   &    43  &  0.767  &   0.818  &  0.750  &  0.783  \\
lgbt             &   157   &    40  &  0.925  &   0.875  &  1.000  &  0.933  \\
migration        &   170   &    43  &  0.767  &   0.708  &  0.850  &  0.773  \\
military         &   312   &    78  &  0.872  &   0.881  &  0.881  &  0.881  \\
misinformation   &    86   &    22  &  0.727  &   0.684  &  1.000  &  0.813  \\
peacekeeping     &   328   &    82  &  0.951  &   0.917  &  1.000  &  0.957  \\
reparations      &   113   &    29  &  0.759  &   0.800  &  0.842  &  0.820  \\
repression       &   312   &    79  &  0.810  &   0.857  &  0.800  &  0.828  \\
technology       &   296   &    75  &  0.707  &   0.744  &  0.744  &  0.744  \\
territory        &    79   &    20  &  0.800  &   0.733  &  1.000  &  0.846  \\
terrorism        &   334   &    84  &  0.655  &   0.655  &  0.783  &  0.713  \\
\end{tabular}
    \caption{Accuracy figures for the document-level \plcon{context} classifier. Note that the validation set is drawn from the same set of annotated articles and is relatively small, meaning that the high accuracy numbers may not be generalizable to real data.}
    \label{tab:context_acc}
\end{table}

\begin{longtable}{lllllll}
\toprule
        \plmod{mode}        &   Train N & Test N&  Accuracy &  Precision &  Recall &     F1  \\ \midrule
        \endfirsthead
        \toprule
        \plmod{mode}         &   Train N & Test N&  Accuracy &  Precision &  Recall &     F1 \\ \midrule
        \endhead
        \hline
        \multicolumn{7}{c}{\textit{Continued}}\\   \bottomrule
        \endfoot
        \bottomrule
        \endlastfoot
        
\vspace{-0.6em} \\
ACCUSE-allege           &     41    &   11  &  0.9091   &  0.9167    & 1.0000  &  0.9565 \\
ACCUSE-disapprove       &     35    &    9  &  0.8889   &  0.9000    & 1.0000  &  0.9474 \\
ACCUSE-investigate      &     36    &    9  &  0.8889   &  0.9000    & 1.0000  &  0.9474 \\
\vspace{-0.6em} \\
ASSAULT-abduct          &     15    &    4  &  0.7500   &  1.0000    & 0.8000  &  0.8889 \\
ASSAULT-assassinate     &      8    &    2  &  0.5000   &  0.6667    & 1.0000  &  0.8000 \\
ASSAULT-beat            &     80    &   20  &  0.9500   &  1.0000    & 0.9167  &  0.9565 \\
ASSAULT-cleansing       &      24   &     7 &   0.8571  &   0.8750   &  1.0000 &   0.9333 \\
ASSAULT-crowd-control   &     12    &    3  &  0.6667   &  0.7500    & 1.0000  &  0.8571 \\
ASSAULT-destroy         &      15   &     4 &   0.7500  &   0.8000   &  1.0000 &   0.8889 \\
ASSAULT-execute         &      17   &     5 &   0.8000  &   0.8333   &  1.0000 &   0.9091 \\
ASSAULT-explosives      &      43   &    11 &   0.8182  &   0.8182   &  1.0000 &   0.9000 \\
ASSAULT-firearms        &      22   &     6 &   0.8333  &   0.8571   &  1.0000 &   0.9231 \\
ASSAULT-heavy-weapons   &      8    &    3  &  0.6667   &  0.7500    & 1.0000  &  0.8571 \\
ASSAULT-primitive       &    118    &   30  &  0.9667   &  0.9444    & 1.0000  &  0.9714 \\
ASSAULT-sexual          &      8    &    3  &  0.6667   &  0.7500    & 1.0000  &  0.8571 \\
ASSAULT-suicide-attack  &     23    &    6  &  0.8333   &  1.0000    & 0.8000  &  0.8889 \\
ASSAULT-torture         &     13    &    4  &  0.7500   &  0.7500    & 1.0000  &  0.8571 \\
\vspace{-0.6em} \\                                                               
COERCE-arrest           &     364   &    91 &   0.9121  &   0.9123   &  0.9455 &   0.9286 \\
COERCE-ban              &      72   &    19 &   0.9474  &   1.0000   &  0.9091 &   0.9524 \\
COERCE-censor           &      63   &    16 &   0.9375  &   0.9000   &  1.0000 &   0.9474 \\
COERCE-curfew           &      65   &    17 &   0.9412  &   0.9000   &  1.0000 &   0.9474 \\
COERCE-deport           &     101   &    26 &   0.9615  &   0.9231   &  1.0000 &   0.9600 \\
COERCE-martial-law      &      81   &    21 &   0.8571  &   0.8235   &  1.0000 &   0.9032 \\
COERCE-misinformation   &     72    &   19  &  0.9474   &  1.0000    & 0.9286  &  0.9630 \\
COERCE-restrict         &     96    &   25  &  0.8800   &  0.8500    & 1.0000  &  0.9189 \\
COERCE-seize            &     99    &   25  &  0.9200   &  0.8667    & 1.0000  &  0.9286 \\
COERCE-withold          &     64    &   16  &  0.9375   &  0.9167    & 1.0000  &  0.9565 \\
\vspace{-0.6em} \\                                                           
CONSULT-multilateral    &      53   &    14 &   0.9286  &   1.0000   &  0.9167 &   0.9565 \\
CONSULT-phone           &     13    &    4  &  0.7500   &  0.8000    & 1.0000  &  0.8889 \\
CONSULT-third-party     &      29   &     8 &   0.8750  &   0.8750   &  1.0000 &   0.9333 \\
CONSULT-visit           &     112   &    29 &   0.8966  &   0.9231   &  0.9600 &   0.9412 \\
\vspace{-0.6em} \\                                                              
MOBILIZE-militia        &     18    &    5  &  0.8000   &  0.8333    & 1.0000  &  0.9091 \\
MOBILIZE-police         &      78   &    20 &   0.8500  &   0.9286   &  0.8667 &   0.8966 \\
MOBILIZE-troops         &     121   &    31 &   0.8387  &   0.8000   &  0.9412 &   0.8649 \\
MOBILIZE-weapons        &      45   &    12 &   0.9167  &   0.8889   &  1.0000 &   0.9412 \\
\vspace{-0.6em} \\                                                           
PROTEST-boycott         &      68   &    18 &   0.8889  &   0.8571   &  1.0000 &   0.9231 \\
PROTEST-demo            &     464   &   117 &   0.8889  &   0.8866   &  0.9773 &   0.9297 \\
PROTEST-hunger          &     109   &    28 &   0.9643  &   0.9375   &  1.0000 &   0.9677 \\
PROTEST-obstruct        &     170   &    43 &   0.9535  &   0.9583   &  0.9583 &   0.9583 \\
PROTEST-riot            &     275   &    69 &   0.8841  &   0.8367   &  1.0000 &   0.9111 \\
PROTEST-strike          &     147   &    37 &   0.9459  &   1.0000   &  0.9200 &   0.9583 \\
\vspace{-0.6em} \\                                                                
REJECT-assist           &     69    &   18  &  0.8889   &  1.0000    & 0.8000  &  0.8889 \\
REJECT-change           &      92   &    23 &   0.6522  &   0.6500   &  0.9286 &   0.7647 \\
REJECT-meet             &      67   &    17 &   0.8824  &   1.0000   &  0.8182 &   0.9000 \\
REJECT-yield            &      64   &    17 &   0.7059  &   0.7000   &  0.7778 &   0.7368 \\
\vspace{-0.6em} \\                                                                
REQUEST-assist          &     248   &    62 &   0.8065  &   0.8095   &  1.0000 &   0.8947 \\
REQUEST-change          &      89   &    23 &   0.6522  &   0.7333   &  0.7333 &   0.7333 \\
REQUEST-meet            &     73    &   19  &  0.8421   &  0.7273    & 1.0000  &  0.8421 \\
REQUEST-yield           &      68   &    18 &   0.8889  &   0.8750   &  0.8750 &   0.8750 \\
\vspace{-0.6em} \\                                                          
RETREAT-access          &      69   &    18 &   0.8333  &   0.8125   &  1.0000 &   0.8966 \\
RETREAT-ceasefire       &      82   &    21 &   0.9524  &   1.0000   &  0.9333 &   0.9655 \\
RETREAT-disarm          &      80   &    20 &   0.9000  &   0.8750   &  1.0000 &   0.9333 \\
RETREAT-release         &     227   &    57 &   0.8947  &   0.8718   &  0.9714 &   0.9189 \\
RETREAT-resign          &      93   &    24 &   0.9583  &   0.9286   &  1.0000 &   0.9630 \\
RETREAT-return          &      20   &     5 &   0.8000  &   0.8333   &  1.0000 &   0.9091 \\
RETREAT-withdraw        &     104   &    27 &   0.9259  &   0.9000   &  1.0000 &   0.9474 \\
\vspace{-0.6em} \\                                                           
SANCTION-convict        &     161   &    41 &   0.9024  &   0.9474   &  0.8571 &   0.9000 \\
SANCTION-discontinue    &     226   &    57 &   0.8246  &   0.8889   &  0.7742 &   0.8276 \\
SANCTION-expel          &     187   &    47 &   0.9149  &   0.9231   &  0.9231 &   0.9231 \\
SANCTION-withdraw       &     152   &    38 &   0.8421  &   0.8400   &  0.9130 &   0.8750 \\
\vspace{-0.6em} \\                                                            
THREATEN-arrest         &      71   &    18 &   0.8333  &   0.8000   &  0.8889 &   0.8421 \\
THREATEN-ban            &      77   &    20 &   0.8500  &   0.9167   &  0.8462 &   0.8800 \\
THREATEN-expel          &      64   &    17 &   0.9412  &   1.0000   &  0.9000 &   0.9474 \\
THREATEN-relations      &      72   &    19 &   0.8947  &   0.8824   &  1.0000 &   0.9375 \\
THREATEN-restrict       &     117   &    30 &   0.7333  &   0.7692   &  0.9091 &   0.8333 \\
THREATEN-territory      &      71   &    18 &   0.8889  &   0.9000   &  0.9000 &   0.9000 \\
THREATEN-violence       &  70    91   &    23 &   0.6957  &   0.8125   &  0.7647 &   0.7879 \\
\bottomrule
\caption{Accuracy figures for the \plmod{mode} classifier. Note that the validation set is drawn from the same set of annotated articles and is relatively small, meaning that the high accuracy numbers may not be generalizable to real data.}
\label{tab:mode_acc}
\end{longtable}

\section{Collecting Training Data with Active Learning and Synthetic Text} \label{sec:annotation}

To date, every automated event coding system has been dependent on human-developed dictionaries operating on text which is either internally parsed by the program (\soft{KEDS, TABARI, ACCENT}) or by an open-source parser (the \soft{PETRARCH} family of coders). These dictionaries require a great deal of skilled human effort to construct, and are language-specific.

In contrast, the various sub-systems of NGEC all ``learn'' from a set of training examples, currently on the order of a couple hundred positive cases, though the system would be improved with more. These can readily be produced by individuals who have only language and domain knowledge, rather than requiring knowledge of the coding programs dictionary specification language (and, typically, some computational linguistics). Example-based coding also ensures continuity of the category definitions over time, which has proven very difficult  across a series of single-propose programs using their own distinct dictionaries. While in the past the process of generating training case has been quite time consuming and tedious, newer approaches such as those incorporated into the web-based \pr ~annotation system\footnote{\soft{https://prodi.gy/}} allow this to be done much more quickly (though without eliminating tedium).

In contrast to the coder-specific dictionaries, annotated cases can be re-used even if the underlying ML/NLP technology changes, as well as being used to test, with minimal effort, new technologies as these become available on commercial cloud platforms or in open source code. Furthermore a large and robust set of annotated cases---which except for cases where the ontology changes can be done cumulatively---allows this to be done in split samples, with some of the cases used to train the model and then the remainder used to test it, which has only rarely been done with the dictionary-based systems, and even then only with small samples. 

Finally, the quality of the annotated cases themselves can be easily assessed to identify clusters of cases where either the annotator was performing poorly, where the nature of the data has changed, typically through the introduction of new vocabulary (e.g. ``ethnic cleansing'' was not used prior to the 1990s), or where we see inconsistent interpretations of the ontology. These issues are common—in fact central—to all contemporary ML projects and consequently a large variety of methods, usually implemented in open source software, have been developed to address them.

One of the key differences between \pc/\pl ~and all of the \cm-based systems is our use of stories\footnote{More precisely,  large fragments of stories: this is discussed in Section \ref{sec:512}.} rather than the single sentences coded in the prior parser-dictionary systems. The clear advantage of this approach is that most of the relevant \plcon{context} of the story is available---though often even full stories, particularly those from local rather than international sources, will assume quite a bit of background knowledge on the part of the reader, such as the names of major cities and roles of important politicians in adjacent countries. The result is that most stories contain multiple events. This can also occur in a single sentence with compound phrases, but this is relatively rare. 

The disadvantage is that the cognitive load on individuals trying to develop training cases is much higher since there is more text involved, and it generally has to actually be read and processed, for example to resolve coreferences. With contemporary coding software such as \soft{\pr } an experienced coder (and the software itself using active learning) can process single sentences very quickly, whereas this is far more difficult (and fatiguing) with full stories.

That said, the overall training required for individuals annotating training cases is dramatically less than that required of individuals developing dictionaries for a parser-based system, and in a system with a mature coding ontology---which we did not have for \pl, as we were developing it as we went along---should require less training, and particularly less training that will seem alien to individuals with a college-level understanding of politics.

\subsection{Collecting Annotations for Contexts}

As noted above, \plcon{context} corresponds to the ``why''-component of a given political event. \pl 's \plcon{context} labels are intended to enable end-users to search for and/or subset events according to specific overarching themes such as environmental-, human rights-, or criminal-related events: Table \ref{tab:context_acc} shows the set we are currently coding. 

In our annotation work, approximately eight expert human annotators labeled whether a particular news story included a given \plcon{context} somewhere in its full text  for future use as training data.\footnote{For budgetary reasons, each article was annotated by a single annotator, which is not the best practice employed by NLP researchers generating evaluation datasets. However, we frequently reviewed annotators' labels as a group and provided feedback on their work. Furthermore, this process is an instance where our protocols changed over time and our training cases aren't completely compatible with our current practice. We originally asked the coders to annotate \plcon{contexts} found \textit{anywhere} in the story, but as we began to deploy the system, we realized that users would generally be able to see only the 512-token text from which the event \plcat{categories} were coded, which could lead to confusion when the coded \plcon{context} was based on texts that could only be accessed from deep within the database. As a consequence we switched to coding only from the 512-token text.} If a news story contained a discussion of a given \plcon{context}, it was considered to be a ``positive case.''  If it did not contain sufficient discussion of a given \plcon{context}, it was considered to be a ``negative case.''  For both the human annotation task and subsequent machine learning classification tasks, news stories were allowed to have more than one \plcon{context}. Allowances were made for ``ignoring'' a news story rather than labeling it as a positive or negative \plcon{context} case.\footnote{This was used sparingly, and primarily for stories that were either too ambiguous for coding, or that were irrelevant to political event data extraction altogether (e.g., shareholder/financial reports for various companies).} The news story sample included articles from the 1990s to the present that were previously employed for ICEWS event extraction.

Individual news stories were supplied to individual annotators in a semi-random fashion via a web-based \pr ~interface.  Annotators were presented with a single news story at a time within this interface, based on a set of \plcon{context}-specific keywords. These keywords were consistently assigned by a single team member for each \plcon{context} category, drawing upon the paragraph-length \plcon{context} definitions included within the Supplemental Appendix to \citet{halterman_et_al2023plover}. The coding interface then updated dynamically to supply increasingly relevant news stories to the human annotator using an ``active learning'' approach. Specifically, as annotators labeled their supplied stories for a given \plcon{context} as positive or negative, additional news stories were identified and supplied to that annotator based not only on the initially supplied keywords but also (and increasingly) using a supervised machine learning model that was dynamically updated. Annotators were shown documents where the model's prediction was least certain, ensuring that each document annotation was maximally informative for training the model. In this manner, and for each \plcon{context}, expert annotators were able to efficiently label the most relevant news stories for \pc 's post-annotation supervised machine learning (data coding) stage.

Actual annotation on the \plcon{context}-specific \pr ~interface required each human annotator to fully read a supplied news story before making an annotation decision. The annotator was then asked to denote whether that story discussed the assigned \plcon{context} category somewhere in the text. The following criteria was also applied. First, a \plcon{context}'s theme need not be explicitly linked to a perceived event within a news story for that news story to receive a positive \plcon{context} annotation. This was to ensure that \plcon{context} annotation was sufficiently de-linked from the other components of the event data annotation and coding process. At the same time, a \plcon{context} needed to be an overarching theme for the story to receive a positive label. Next, news stories can have multiple \plcon{contexts} and that the presence of one \plcon{context} should not preclude the labeling of that story for an additional \plcon{context} where applicable. Finally, emphasis should be placed on general thematic discussions---as opposed to mentions of specific actors and actions---when annotating \plcon{contexts}. This ensured that the subsequent annotation of entities, \texttt{event types}, and \plmod{modes} were sufficiently de-linked from the \plcon{context} component.

Annotators were asked to label at least 250 positive cases (i.e., news stories) for each of \pl 's 37 general \plcon{context} categories\footnote{This typically ensured that the number of negative labels significantly exceeded these positive cases for each \plcon{context} category.}  or in the case of 13 rarer \plcon{context} categories, to label a total 1,000 total positive or negative \plcon{context} cases.\footnote{This ensured between 75 and 200 positive labels and an additional 800-plus negative labels for these rarer cases.}.

\subsection{Collecting Annotations for Event \plcat{categories}}

\texttt{Event type} annotation followed a similar annotation process as \plcon{context} annotation on \pr, using the same annotators as above. However, unlike \plcon{context} annotation, event \plcat{category} annotation entailed the annotation of a specific subset (i.e., span) of a news story's text rather than solely annotating a news story’s entire text. The goals of event \plcat{category} annotation were twofold. First, and for a given \plcat{category}, we annotated whether a particular news story contained a record of an event of a particular type. Crucially, and unlike previous automated event coding systems, annotators worked at the level of the document rather than the sentence, in order to handle the common situation of a single event being reported across multiple sentences. Instances (i.e., stories) where an event of a certain type were present were considered ``positive cases'' of that \plcat{category} whereas instances (i.e., stories) where an event \plcat{category} was absent were considered ``negative cases.''  As was the case for \plcon{context} annotation, allowances were also made for skipping news stories and not assigning a label where appropriate.

Second, for stories that contained an event of a particular type, annotators highlighted the specific passage(s) of the news story text that described that event (in terms of who did what to whom, and where/when). We then leveraged these annotations via transformer-based neural networks to detect specific event spans within un-labeled news story texts and determine the event \plcat{category} of those detected event spans across \pc 's full news corpus.\footnote{We are currently not using the span-level information, but plan to use it in the future to identify multiple instances of the same event type and to better guide the QA/attribute identification step.} 

Within the \pr ~event \plcat{category} interface, annotators annotated one event category at a time and sought to annotate at least 250 positive cases in each instance.\footnote{Except for \plcat{THREATEN}, see Table \ref{tab:category_acc} and the discussion in section \ref{subsec:editing-agree}.} Annotators achieved this positive-label threshold for all 16 event \plcat{categories}, which generally ensured 1,000 or more comparable negative labels for each event \plcat{category}. Annotators also labeled instances where an event was present even if the specific actor and/or recipient of that event was ambiguous. Similar to \plcon{context} annotation, this ensured the annotation of a \plcat{category} (and its overarching presence or absence) was de-linked from the broader process's attribute annotation components. It was important at this stage that annotators ensure their (non)identification of an event in a news story was independent of their (in)ability to identify an event's actors and recipients.

\subsection{Collecting Annotations for Modes}

\plmod{mode} annotation followed a similar \pr -based approach to the earlier \texttt{event type} and \plcon{context} annotation activities described above. In this case, the same team of expert annotators annotated news stories for the presence or absence of individual \plmod{modes}. This particular annotation process was distinct from the earlier event \plcat{category} and \plcon{context} annotation processes in that \plmod{mode} annotation required a priori that the individual news stories being annotated were already determined in the previous phase to contain a particular \plcat{category}. To ensure a sufficient sample size of such cases, synthetic stories were also generated for inclusion within each relevant \texttt{event type}'s \plmod{mode} annotation sample (see \citet{halterman2022synthetic} for details). For this combined set of real and synthetic stories, \plmod{mode} annotators labeled each for the presence (a ``positive case'') or absence (a ``negative case'') of each individual \plmod{mode} associated with that \plcat{category}.

Each annotator was assigned one event \plcat{category} out of the 11 that have defined  \plmod{modes} in \pl . Annotators reviewed the relevant \plmod{modes} for their assigned \plcat{category} within the \pl ~manual and were then supplied a series of news stories associated with their assigned \plcat{category}. For each supplied news story, a specific single \plmod{mode} for evaluation was provided via a prompt at the top of the \pr ~\plmod{mode} annotation interface for a given annotator's consideration. Based upon this supplied information, annotators were instructed to annotate whether the first  relevant event within a supplied news story contained evidence of a particular \plmod{mode}. During this process we determined that some event \plcat{categories} with an exceptionally large number of \plmod{modes} ultimately required more (at most two or three) annotators. Annotators annotated at least 50 positive cases (i.e., news stories) for each event \plmod{mode}, with no specified threshold of negatively labeled cases. Ultimately, this target was reached for all relevant \plmod{modes}.  Differences in the frequency of some \plmod{modes} relative to others for the same  \plcat{category} often ensured that \texttt{each event} type had at least some \plmod{modes} that received over 200 positive labels.

\subsection{Collecting Annotations for Attributes}

Attribute annotation---relating to an event's actor, recipient, location, and time---was similarly completed using the same \pr ~interface and coders described above. In this case, such annotations were predicated upon previously identified events (and hence real and synthetic news stories), and annotators were asked to highlight a selected event's relevant actor, recipient, location, and time of occurrence; as well as instances where the same actor and/or recipient were mentioned multiple times in different manners throughout a chosen annotation story. This was implemented for all 16 \pl ~event categories in a manner that ensured that each received at least 100 entity-annotated stories. Further annotation continued throughout the development of the pipeline to ensure enough data to train and evaluate the QA model.

\subsection{Refining the training cases using automated methods}

In contrast to dictionary-based methods, training cases can be easily assessed for internal consistency by simply running the models---or, ideally, an ensemble of models, ideally using multiple methods---against the training data and identifying any cases that are consistently tagged as false positives (that is, cases coded as negatives that code as positives) or false negatives, then eliminating these. With human review, this has the additional advantage of identifying potential ambiguities in the category definition, which is rarely obvious from dictionaries, and performance issues for individual coders (a standard protocol for well-resourced projects but sometimes difficult if you are pleading for volunteer assistance).

Software---both custom or open-source (see, for example, CleanLab \citep{northcutt2021confident})---for doing this is straightforward, and we've already run our cases through one or two cycles of this, and will be doing more as the system develops. We would note that it is \textit{possible} to do this as a fully-automated iterated process but this will produce a lock-in effect---the system will become increasingly focused on the cases which dominated the true-positives and true-negatives in the initial stages---and these may not necessarily correspond to the best interpretation of the coding framework, and in fact atypical ``corner cases'' may be where most of the additional human annotation work needs to be invested: an example of this would be differentiating street protests from diplomatic and legislative protests in the \plcat{PROTEST} category. Ideally this refinement would be done against a large robust validation set, as discussed in Section \ref{sec:bench}, and were that available, the system could be fully automated, but at present we don't have this.

\subsection{Developing cases for rare event types: synthetic cases}

A problem we quickly ran into on some of the categories, whether event, mode, or context, was rare event types, where the yield across our corpus of potential training cases was below 5\%, and sometimes well below 1\%. This situation is problematic both from the perspective of the sheer labor required to find sufficient training cases, and also tends to be quite fatiguing and demoralizing for the human annotators, who need to read through dozens of irrelevant cases before occasionally finding a positive case.

We tried addressing this with two methods, neither wholly satisfactory. First, because we were doing this work as part of a U.S. government project, we had access to the licensed  news stories texts that were coded for the positive cases (which cannot be provided in the public ICEWS dataset due to licensing constraints). We have millions of these texts covering a 20-year period, and in addition, for many of the event and \plmod{mode} categories we are coding, there is a straightforward translation between a detailed \cm ~code and a \pl ~event-mode-\plcon{context} triple; this translation table, which we've used widely in this project (and is based on early \pl ~work funded by NSF), is available on our Github site.

Unfortunately, this proved to be only a partial solution. First, the \soft{iData} texts coded into \cm ~are, of course, only single sentences, whereas for training cases we needed complete stories. In addition, while we have access to \soft{iData}, we do not have access to the complete original Factiva and other feeds that went into this, and even within \soft{iData}, we only have texts that generated positive classifications. Second, quite a few of the new \pl ~\plmod{modes} and \plcon{contexts} do not have unambiguous \cm ~equivalents---this applies to almost all of the contexts---and so we had nothing to work with. So in the end, texts extracted iteratively from \soft{iData} using SVM-based classifiers were useful for generating annotation sets that had significantly higher yields, thus reducing the cognitive burden on the annotators, but were not sufficient by themselves for producing useful training cases.

A second approach we tried was using  GPT-2 \citep{radford2019automated} to generate synthetic cases, particularly for those \plcon{contexts} where we had no \cm/\soft{iData} equivalents. Again, this provided some useful cases but was only a partial solution due to two limitations of the technology. First, we could typically get credible synthetic text out to about two sentences, well below the 512-token level of actual stories. Second, the text generators tended to lock into a small number representative verb phrases---for some reason, for example, it liked to generate \plcat{ASSAULT} cases where protesters attacked police with baseball bats---while simply varying the actors, and after a few instances these added no new information to the training set, and would in fact distort it. However, optimizing GPT-2's generation hyperparameters to improve the quality of generated text or using GPT-3 (and successors) are promising options for future annotation work \citep{halterman2022synthetic}.

Despite these limitations, we have incorporated some synthetic cases: These have the advantage of being easily reviewed with a reasonably high yield (25\%+) and, very importantly, have zero intellectual property constraints. With current (and future) LLM technology---\cg ~and its rapidly proliferating derivatives---we would probably be able to get far better results, and particular easily get stories compatible with the 512-token framework, and resources permitting we will undoubtedly explore this in the future.

\subsection{Caveats around the annotated data}

In interests of transparency, we would note that the existing training cases leave a lot to be desired on at least three dimensions:

First, because this was done on a low budget and very short initial time frame compared to the magnitude of the task, and was done without the low-cost skilled labor available in university research environments, we did not employ the standard large-scale coding project protocols of first assembling and thoroughly training a group of coders, doing multiple initial coding experiments to refine the codebook before producing the data we would actually use in the modeling (we did some of this), via these initial steps determine which coders were reasonably proficient (conveniently, those who aren't very good usually just quit), and doing final coding over a period of months, under close supervision, and then finally, once all of this preliminary work had been completed, and if needed, re-coding cases where ambiguities were found. That was the methodology employed in the \soft{KEDS/TABARI} work using student coders and it occurred over a period of about fifteen years, and in the BBN/\soft{ACCENT} work where it occurred over about five years using professional coders.  We had only about four to six months of part-time labor.

Second, while the \pl ~event categories were reasonably stable going into the project, the \plmod{modes} and \plcon{contexts} were not, particularly the \plcon{contexts}. Furthermore, while the \plmod{modes} for the most part corresponded to \cm ~subcategories, there is little in \cm ~corresponding to the \plcon{contexts}, so we had little help from either the \cm ~data or \soft{PETRARCH} dictionaries implementing \cm. 

Third, this was further complicated by the fact that some of the \plmod{modes} and \plcon{contexts}, while of interest to our sponsors, are extremely rare in the datasets we were working with due to these having previously been filtered by \cm/\soft{ACCENT} coding (that is, we didn't have unfiltered news feeds).\footnote{There's an interesting parallel here during the development of \cm ~at the University of Kansas during 2000-2003: we decided that the codebook would contain at least three annotated sentences from actual Reuters news reports for each code, and ended up dropping several when we couldn't locate sufficient unambiguous examples in our roughly 20 years of news reports from the Middle East.} Synthetic cases were a partial solution to this issue---and in the future, we will almost certainly use \soft{ChatGPT} and its successors, which are far superior to the methods we had available a year ago---but only partial.

None of these problems have prevented us from implementing what we have found to be a credible system, but were we to do this over, either with professional coders or with well-trained undergraduate and graduate coders in a university environment, and with a longer time frame, standard coder training protocols would definitely be practical. On the positive side, erroneous codings in example-based systems are far easier to detect than problematic phrases in dictionary-based systems, and as time and resources become available, we expect to be doing this.

\section{Pre- and post-filtering}

\begin{itemize}
    \item Within a text we remove as assortment of datelines,\footnote{Datelines, which, alas, do not have a standard format across sources, are particularly problematic for the subsystems trying to identify locations, as the dateline is typically in a major city that may be hundreds or thousands of miles from the actual event. In a presumably apocryphal story beloved by Africanists, in the 1960s an editor at the \textit{Times of London} telegraphed their branch in Cape Town asking them to cover some unrest in West Africa and received the reply ``You cover it: you're much closer.'' The ICEWS system, for what it's worth, does a really lousy job on this task since location-resolution was added as an afterthought.} attribution/authorship, story components (e.g. \texttt{Section:, PHOTO:}), editorial processing notes (e.g. \texttt{Embargoed until\ldots, Corrected spelling of\ldots}), and other repeated boilerplate text
    
    \item Stories that are substantially longer than 512 words: these only rarely report contemporary events and more likely are background or historical discussions, or transcriptions of speeches
    \item Stories that are very short: these typically occur when a story has been improperly parsed at some point so only the headline, or a fragment of the headline, remains
    \item Stories consisting mostly of numbers, which are usually market reports or sports scores. You'd think Factiva could filter these out but for some reason this is imperfect
    \item Stories dealing with crimes that have no political content, typically murders or unpleasantness occurring to tourists. We had assumed Factiva had already gotten rid of these; we assumed wrong, and these originally played havoc with our \plcat{COERCE} and \plcat{ASSAULT} codings.

    What of crimes that may or may not be political, such as corruption and attacks (or killings) that may or may not be politically motivated? There's a decided grey areas here that is further complicated by false accusations, with accusations of various forms of sexual misconduct, the particular modes of what is considered unacceptable varying widely in various cultures, being among the more popular. And sometimes a robbery is just a robbery, so in the end there will be some unresolvable cases here.
    
    \item Stories dealing with accidents and natural disasters, which we aren't coding. These are less problematic than crime stories in terms of false-positive codings, but still use up resources. They also have a grey-area component in that major disasters tend to generate international \plcat{AID} responses.

    \item We exclude events that are wholly US domestic events. 

    \item \textit{Sentences} that involve a negative, for example a politician saying that they would \textit{not} do something. These form a curious set of cases since due to the human proclivity for over-estimating novelty, such constructions \textit{seem} as though they would be common and important. In fact, they are quite rare as far as \textit{events} are concerned---negative \textit{verbs} such as ``reject'' or ``accuse'' are used, but not simple negation---and most commonly occur when someone is talking about the supposed position of an \textit{opponent} (in other words, using the novelty of negation as a rhetorical technique). Detecting such sentences was potentially a major issue---and in fact negation is a major technical challenge within the field of computational linguistics---but conveniently the \soft{spaCy} NLP system does this for us and in the pipeline it involves only a couple lines of code.

    \item ``Composite stories'' that are lists of headlines, major stories, upcoming events and the like were quite useful in single-sentence classifiers, since a single story would provide information on multiple discrete events. Unfortunately, they are problematic in full-story coding because they will code to a large number of possibly unrelated events, and they are seriously problematic as training cases since they provide an incredibly diffuse signal that is exactly the last thing one wants in a training case.\footnote{This is a place where we failed to properly instruct our human annotators to not include these, and in some early versions of the training sets they were causing considerable problems. Most are gone now, but not all due to the non-standardization issues.} The major international sources such as Reuters and BBC have fairly standard formats for these and they can be readily eliminated; this is less true for various national-level sources or other sources. These are also easily missed as one becomes fatigued doing human annotation: the first 2/3rds of a story will look perfectly normal, but then one can miss that the remaining third contains a couple unrelated headlines or summaries.
\end{itemize}

\subsection{Calibration and post-classification filtering}\label{sec:calib}

As noted above, a key advantage of machine-learning classification approaches in contrast to dictionary-based pattern matching is that we get a real-valued classification score rather than a binary true/false determination. In principle, the classification score---which is an arbitrary metric, not a probability---is proportional to the confidence of classification. Consequently once we had determined a set of consensus models, we ran these over a large representative sample and determined the distribution of the positive classification scores,\footnote{We experimented with a couple other metrics like the difference and ratio of the positive and negative scores, but these didn't seem to provide any obvious added advantages} then experimented with setting the cut-off points are various levels---it was obvious from the beginning these needed to be relatively high, e.g. in the 70 to 95 percentile range---and then [quickly] reviewing the output to assess the degree to which we had eliminated cases which, as humans, did not appear to belong.

Our assumption, of course, that the higher we set the threshold, the better we'd find the results. Alas, this didn't really prove to be the case: by definition, of course, we got fewer cases as we raised the percentile limit, but the ratio of those we, as humans, thought belonged barely changed. This is because the machine was simply implementing, with ever-greater discretion, what it had inferred from the [decidedly imperfect] training cases, not by grokking our intuition as to what belonged in each \pl ~category.

On the one hand, one could interpret this as simply the consequence of inadequacies of our training cases but, on further reflection, we think it may go beyond this. For starters, in many (or most?) machine learning methods, raising classification threshold does in fact trim the output to what is in some sense the centroid of the training cases, which is to say, only the best fits. But that didn't happen, and the reason may be the nature of LLMs: the vast bulk of the knowledge incorporated into any LLM is in the unmodified parent model, in our case \soft{distillBERT}, which is looking for general language characteristics, and our quite limited training cases---perhaps 500,000 words at best, against the 3.4-billion words used to train \soft{BERT}---only refine this. 

But more generally, these models are quite capable of picking up regularities which, while very real in some vast LLM vector space, we can't see as humans, particularly humans who desperately want this thing to work, sooner rather than later. 

A good example of this occurred when we were trying to refine the classifier for the \plcon{gender} context.\footnote{This is actually an SVM but the possible issues with vectorization based on LLMs may still hold.} This already starts out as a relatively difficult case: the \plcon{context} is reasonably clear for our human annotators, but relevant events proved to be \textit{very} rare in the available news records, particularly after criminal cases of sexual assault were removed. As a consequence, the system vastly over-classified, and it showed a particular affinity for events involving major powers in Asia. 

But as is well-known, until quite recently ICEWS hugely over-sampled events in Asia, since Asia and the Pacific was the focus of the original DARPA-sponsored ICEWS project, which was reflected in the corpus available for selecting training cases. When a classifier trained on those cases is then applied to the \textit{current} corpus, where the news search terms have been modified to reduce the Asian selection bias, when the system ``sees'' an Asian case, it ``looks'' similar to the gender cases, which are otherwise quite amorphous due to the low number of training cases, and so it figures the \plcon{context} must be \plcon{gender}.  We've not had the time or resources to formally test this hypothesis, but it seems at least plausible.

So how to get the keywords, particularly as we were under a tight deadline. In the case of the events, we started with the primary verb lists from the \soft{PETRARCH} dictionaries, which are open source, and then used the \soft{spaCy word2vec} similarity measures against frequent verbs in the positive training cases (after the usual elimination of stopwords and the like), followed by some additional tweaks. The process for the \plcon{contexts} (\plmod{modes} were already working sufficiently well) was similar except our baseline were the \plcon{context} titles and representative words from the \plcon{context} descriptions in the \pl ~manual.\footnote{Okay, in fact we didn't think about using the manual until the end of the process but that is how we should have done this. We also should have worked with bigrams, not just single words--\soft{spaCy}'s similarity function is fine with this---but ran out of time: go ye and do otherwise.}. The final step for \plcon{contexts} involved an interactive process using an program where several of us thought of various candidate terms which would be tested against a representative corpus to see whether these actually were present---this feedback occurred in seconds---and converged on a human-curated list.

This final step dramatically, and systematically, reduced the number of positives \plcat{category} and \plcon{context} assignments,\footnote{Occasionally, if all of the \plcat{category} assignments are filtered out, an entire story is removed, but more commonly a story goes from having, say, eight \plcat{categories} assigned to only two \plcat{categories}. Provided it has a valid \plcat{category}, stories are retained even if they have no \plcon{contexts}: this is not uncommon as the \plcon{contexts} are no intended to be exhaustive} with two effects. In the case of primary events, it changes the event distribution, which originally had been more or less uniform (that is, roughly equal events in each category, presumably a consequence of our consistent use of 1:1 positive:negative ratios in the training sets) to something much closer to the distribution of events in ICEWS, the only thing we have that is close to a reference set. This appears to have also generally occurred for the \plcon{contexts} in the sense that the post-filtering eliminated large numbers of assignments \plcon{contexts} we believe to be relatively rare, but few in \plcon{contexts} such as \plcon{military} which we know to be common.

\subsection{Comparative assessment of effect of filtering and human editing for  \plcat{AGREE, CONSULT, REJECT, THREATEN}, and \plcat{ASSAULT}}\label{subsec:editing-agree}

Late in 2022 we made two modifications to the event \plcat{category} classification: several of the filters discussed in Section \ref{sec:preprocess} and an intensive manual review of the entire training corpus for three of the most frequently coded categories in ICEWS (or specifically, their \pl equivalents) \plcat{AGREE, CONSULT}, and \plcat{ASSAULT} and two categories, \plcat{REJECT} and \plcat{THREATEN} where the split-sample accuracies were lower than our target of 0.80. Table \ref{tab:category_acc} shows an initial comparison of split-sample statistics in each of the category, where ``Mk7'' refers to the older process and ``Mk9'' to the newer. Note that these a \textit{split-sample}, not tests against an independent validation set (see Section \ref{sec:bench}), which is particularly important for ASSAULT. This is based on 8 estimated models (and hence 8 different split samples) for each category: this takes about half an hour to run in Google's Colab GPU environment.

For starters, note that all of the human review reduced the total size of the training size by around 30\%. Also note that the number of training cases in \plcat{THREATEN} is very small in both Mk7 and Mk9---as noted below, our Factiva sample came from a period of relatively low conflict, so it was difficult to find examples of this behavior---which probably explains the low accuracy. The lower accuracy for \plcat{REJECT} remains something of a mystery, since the sample size itself is large, but the mix of examples apparently remains difficult for the LLM to characterize.

For the most part, the results, particularly on the F1 metric (a balance of precision and recall) are a wash: it goes up a fair amount in \plcat{AGREE} and \plcat{THREATEN}, down a bit in \plcat{REJECT}, and is unchanged in \plcat{CONSULT}. So far, there doesn't seem to be a consistent pattern on how precision vs recall change. Accuracy (``acc'') generally drops between Mk7 and Mk9, but this is less important since the samples here are artificially balanced (50/50 mix of positives and negatives) in comparison to the actual data.

So, good or bad, and is this labor-intensive review worth doing? Probably ``good'' in the sense that the filters eliminated some categories of things we don't want to be coding that may have been artificially increasing the various metrics because they were distinctive and the systems more or less holding steady without these. Also performance in the split-samples is holding steady with the smaller training sets.

The outlier is  \plcat{ASSAULT}, where the F1 drops by more than half (0.73 to 0.34) albeit with the accuracy dropping only by about 15\%, and the reason is the huge drop in recall, by about 2/3rds (0.77 to 0.28) (``recall'' you will recall, is the proportion of positive test cases that are correctly identified). This major difference is almost certainly due in large part to the removal of the street crime cases, of which there were many (note that the number of training cases is reduced by almost 50\%), and which are generally relatively easy to code, so this inflates the metrics in Mk7.  Meanwhile we obviously don't have enough ``conventional'' \plcat{ASSAULT} cases, probably because our available training cases were sampled from a relatively quiet period, after most of the violence in Syria, Iraq, and Afghanistan, and before Ukraine-Russia: if additional work is done on the system all of these crises can provide training cases for \plcat{ASSAULT}, and these will likely improve the performance of the system over the 2010-2022 period, which includes quite a number of extended large-scale conflicts.

This might also be a useful place to discuss the limitations of human-coded event data which is, to put things mildly, anything but perfect. For example, \citet{douglass_et_al2022ICBe} introduces the new ICBe dataset which is based on the Breecher-Wilkenfeld International Crisis Behavior data. ICBe has an absolutely ideal set of texts to code—thoroughly edited retrospective summaries written in English by experts---and still their average intercoder agreement (granted, in a very complex ontology) barely passes 60\%, and their experts are down around 55\%, and the only place they pass 80\% is agreement on whether an event exists at all.\footnote{The intercoder reliability figures were present in V1 of the paper on arXiv but are absent in v2.} Those numbers are also consistent with a study published about ten years ago on one of the Uppsala Conflict Data Project data sets, and UCDP uses professional coders as well as doing retrospective (one-year lag) coding. In our internal tests on our annotated cases,  our split-sample accuracies are generally above 80\%, and frequently well into the 90\% range.\footnote{In a future version of the paper, we will provide these metrics for all of the categories: through lack of foresight, we didn't retain these for the most recent estimation runs} Consequently for most \pl ~event categories NGEC is doing well above human performance, as well as being able to maintain global data set in near-real-time, which is simply impossible with human coding.

\section{Glossary}

\begin{longtable}[]{p{3cm}p{12cm}}
     \toprule
        & Glossary and acronyms\\ \midrule
        \endfirsthead
        \toprule
        & Glossary and acronyms\\ \midrule
        \endhead
        \hline
        & \textit{Continued\ldots}\\   \bottomrule
        \endfoot
        \bottomrule
        \endlastfoot
        

\soft{ACCENT} 	& The \cm ~coding system developed by a BBN team which took over the ICEWS data production from Lockheed around 2012; it is derived from a general-purpose BBN NLP system named SERIF\\

AI &	artificial intelligence, an earlier term for ML\\

BBN	& A large defense contractor with expertise in signal processing, among many things, and briefly responsble for the production of ICEWS data\\

\soft{BERT}	& A very large neural network transformer model developed and made available by Google. \\

\cm ~&	An event data ontology originally developed under NSF funding in the early 2000s for academic research, then later adopted by the DARPA ICEWS project and subsequent research\\

\cg & An LLM developed by OpenAI currently (Feb-2023) taking the technorati world by storm \\

DARPA	& United States Department of Defense Defense Advanced Research Projects Agency\\

\soft{distilBERT} & A smaller, more efficient subset of the BERT LLM\\

Factiva	& A major data aggregator which has provided the source stories for ICEWS/iData \\

\soft{Github} & A widely-used cloud-based repository for open source software \\

GPU	& Graphical processing unit, a specialized but widely available chip that speeds neural network processing by a factor of 30x to 60x\\

\soft{Geonames} & An open access global geographical database containing over eleven million placenames\\

\soft{GPT-3} & An LLM that forms some of the base of \cg\\

\soft{HuggingFace} & An AI company which has developed and/or hosts a large number of open source transformer software and models\\

ICEWS 	& Integrated Conflict Early Warning System, both the name of a major DARPA conflict forecasting project (2008-2011) focused on Asia, and the globally-focused dataset subsequently developed from the technology which is currently updated weekly on the open-access Dataverse site\\

\soft{iData} & The government version of the ICEWS data produced by PITF using the Lockheed and BBN software\\
JSON	& a very general and now widely-used data-interchange format \url{https://www.json.org/json-en.html} \\
\soft{KEDS}	& Kansas Event Data System, the first automated coder producing data which resulted, by the early 1990s, in refereed academic publications\\

Leidos & A large U.S. defense contractor currently responsible for the production of the ICEWS and iData event data sets\\

LLM & Large language models: very large neural-network-based models trained on vast amounts of text\\
ML 	& machine learning\\

Lockheed & A large U.S. defense contractor whose systems won the original DARPA ICEWS competition and consequently for several years was involved in the production of the ICEWS datasets\\

NER	& named entity resolution/recognition, an NLP problem\\

NGEC	& New generation event coder, the working name for this project\\

NLP	& natural language processing\\

NSF	& National Science Foundation\\

\soft{PETRARCH}	& A set of experimental event coders developed in the mid-2010s using the open Stanford CoreNLP language parser and the TABARI event and actor dictionaries\\


PITF	& Political Instability Task Force, a long-running (since 1995) multi-agency conflict analysis project which took over ICEWS around 2020\\

\pl	& The event data ontology we are using to replace \cm\\

\pr & A program from Explosion AI for rapid annotation of text data\\

Python & A general-purpose programming language widely used for NLP work and data analytics more generally. Also a seriously invasive reptile in the Florida Everglades\\

QA	& Question-answering, an ML/NLP technique we use for entity identification. Not to be confused with ``quality assurance'', though we also need to be doing that.\\

\soft{roBERTa} & A smaller, more efficient subset of the BERT LLM\\

\soft{spaCy} & A extensive open source NLP software suite produced by Explosion.AI. \url{https://spacy.io/}\\

SVM	& Support vector machines, a robust and widely-used ML method for classification. It is a family of algorithms working on data in very high dimensional spaces, not a physical machine. \url{https://en.wikipedia.org/wiki/Support-vector_machine}\\

\soft{TABARI}	& An event coder written in the C++ language developed under NSF funding in the early 2000s and used by the Lockheed team to win the DARPA ICEWS forecasting competition\\

Wikipedia & An open source  online encyclopedia, created and edited by volunteers around the world and hosted by the Wikimedia Foundation.\\

\bottomrule
    \caption{Glossary and list of acronyms.}
    \label{tab:glossary}
\end{longtable}

\end{appendix}
\end{document}